\documentclass{article} 
\usepackage{iclr2027_conference,times}

\usepackage[utf8]{inputenc}
\usepackage[T1]{fontenc}
\usepackage{textcomp}

\usepackage{amsmath}
\usepackage{amssymb}

\usepackage{graphicx}
\usepackage{booktabs}
\usepackage{multirow}
\usepackage{array}
\usepackage{adjustbox}
\usepackage{placeins}
\usepackage{newunicodechar}
\newunicodechar{–}{\textendash}

\newcommand{\statpair}[2]{\mbox{#1{\scriptsize\,$\pm$\,#2}}}

\usepackage{float}

\usepackage[hidelinks]{hyperref}
\usepackage{url}

\title{StableGrasp: Reconstructing Physically Stable Human Hand Grasps from Single Images}

\author{Han Jiang, Etienne Vouga, Qixing Huang \& Georgios Pavlakos \\
Department of Computer Science \\
The University of Texas at Austin \\
Austin, TX 78712, USA \\
\texttt{han.jiang@utexas.edu, evouga@cs.utexas.edu} \\
\texttt{huangqx@utexas.edu, pavlakos@utexas.edu}
}

\iclrfinalcopy

\begin{document}

\maketitle

\begin{figure}[H]
\input{sections/teaser}
\end{figure}

\begin{abstract}
Reconstructing a physically stable human grasp from a single RGB image is challenging because physically modeling grasps is itself difficult, and the problem requires estimating not only a visually constrained hand pose but also a control target that stabilizes the grasp.
Existing methods either model only visual hand geometry without considering physics, or rely on less plausible physical modeling, which limits the physical validity of the resulting grasps.
In this paper, we present StableGrasp, a differentiable simulation-based optimization framework that explicitly separates the visual hand pose from the control target that determines the grasping forces.
Our method jointly optimizes hand geometry and control by minimizing the kinetic energy of the grasp in a differentiable simulator, while regularizing the hand geometry to preserve visual consistency and geometric plausibility.
The reconstructed grasps are substantially more stable under rigorous physical simulation, while remaining visually consistent with the input images and geometrically plausible.
Experiments show that our approach produces far more stable grasps than alternative hand-control strategies, benefiting visual-only grasp reconstruction pipelines by turning their outputs into physically stable grasps.
\end{abstract}

\section{Introduction}
\label{sec:intro}

Understanding how humans grasp objects is a fundamental problem at the intersection of embodied intelligence, computer vision, and computer graphics. Accurate reconstruction of hand-object grasps from images is needed for many applications including digital human modeling, AR/VR, robot learning from human demonstrations, and general understanding of dexterous manipulation.

Existing hand-object reconstruction methods have achieved impressive progress in recovering visually plausible grasps from images. Both optimization-based approaches~\citep{easyhoi} and feed-forward models~\citep{HORT,hamer,WiLoR,unihope} can now produce high-quality 3D hand poses with reasonable hand-object spatial relationships that are consistent with the input image. However, most of these methods stop at visual fidelity: the poses are evaluated in terms of reconstruction accuracy, pose consistency, and/or geometric plausibility (i.e. the hand is in contact with, but does not penetrate, the object). A grasp may look convincing when rendered in 3D while still being impossible to execute or physically unstable in reality.

More recent work (surveyed below in Section~\ref{sec:relatedwork}) recognizes the importance of physics in grasp reconstruction and synthesis. A common approach is to analyze stability of the grasped object in terms of force closure, which characterizes whether frictional contact forces at a static set of contact points could, in principle, hold the object in place against gravity. Such analysis does not address how the hand should actually be posed and actuated, and the corresponding grasp may still be impossible if the hand cannot produce and maintain the required contact forces.

We argue that stable grasp reconstruction is fundamentally a control problem, and cannot be treated as a visual or geometric optimization task divorced from inference of joint torques that hold the object stably in place over time. For example, a hand may appear to pinch a cup, yet if the pinch force is insufficient or unevenly distributed, the cup will slip or fall rather than remain firmly put. We therefore present StableGrasp, a method for monocular stable human grasp reconstruction that evaluates stability from the viewpoint of the coupled dynamics of the hand and object under simulation. StableGrasp also achieves strong visual stability: the reconstructed pose closely matches the image observations, and during simulation, the reconstructed grasp remains nearly motionless.
To the best of our knowledge, StableGrasp is the first method to pursue this combination of rigorous physical stability and visual fidelity.

We evaluate our method on several established datasets (\textbf{DexYCB} \citep{dexycb}, \textbf{HO3D} \citep{ho3d}, and \textbf{CORe50} \citep{core50}) against baselines such as heuristic~\citep{dexgraspanything} and learning-based~\citep{dgrasp} hand control. Our results show that compared to these baseline methods, StableGrasp substantially improves physical stability while producing reconstructed grasps that are visually consistent with the images and geometrically plausible.

In summary:
\begin{itemize}
    \item we formulate monocular stable human grasp reconstruction with a notion of physical stability defined by forward simulation of the full hand-object system;

    \item we propose a two-phase physics-based refinement framework that introduces a kinetic-energy-driven optimization for stable grasp reconstruction, while preserving the visual and geometric fidelity of the reconstructed grasp;

    \item we use this framework to build a complete monocular reconstruction pipeline that recovers both the visual grasp configuration and a control target whose induced joint torques stabilize the grasp in physical simulation.

\end{itemize}

\section{Related Work}
\label{sec:relatedwork}
\paragraph{Human Hand Grasp Reconstruction}
Recovering the 3D pose of a human hand and the grasped object from image observations has been extensively studied in computer vision.
Methods for hand-only reconstruction estimate the pose and shape of the 3D hand~\citep{boukhayma2019hand,ge2019graph,metro,meshgraphormer,hamer,WiLoR}. Methods for hand-object reconstruction also estimate the geometry of the object and the spatial relationship between hand and object. Methods attacking this problem have proposed using interaction constraints~\citep{tekin2019ho,hasson2019obman}, contact-aware refinement~\citep{contactopt}, implicit hand-object representations~\citep{alignsdf,gsdf,hoisdf}, unified hand-object estimation~\citep{unihope, easyhoi}, or hand-guided reconstruction of held object geometry~\citep{ihoi,ddfho,chord,dsco,followmyhold,HORT,mccho}. Many of these methods represent the hand using MANO~\citep{mano}, a parametric model that maps the pose and shape of the hand to a skinned mesh. 

However, these methods only aim to recover visually-plausible hand and object poses that are consistent with the input observations, while StableGrasp's reconstruction is physically stable, verified by forward simulation of the coupled hand-object system.

\paragraph{Physics-aware Grasping Assuming Static Hands}
The stricter problem of recovering a grasp that is stable under gravity has also received significant attention. ``Physical stability'' has been modeled in several different ways. One simplified approach is to consider only \emph{one-way} hand\textrightarrow{} object coupling, and to analyze stability of the object in terms of contact forces exerted at a set of contact points where the object touches a static hand.

Classical grasping theory in robotics studies this problem through force closure and grasp wrench space analysis~\citep{ferrari1992planning,bicchi2000robotic}, which evaluates whether contact forces can resist external object wrenches. This and similar quasi-static criteria have been widely used for dexterous grasp synthesis and evaluation~\citep{dfce,graspqp,dexgraspnet,dexgraspnet2,taskdexgrasp,frogger}. Along similar lines, works like Grasp'D~\citep{graspd} and Fast-Grasp'D~\citep{fastgraspd} optimize hand configurations with simulation-based stability objectives, while DeepSimHO~\citep{deepsimho} and UGG~\citep{ugg} use learned simulation surrogates to refine reconstructed interactions. Other methods introduce explicit contact forces at fingertips as optimization variables~\citep{physicalinteraction,physicsawarehoi} or enforce physics-inspired priors on sliding and penetration at the contact points~\citep{handypriors}.

With a static hand, object stability does not establish whether an actuated hand can generate the required contact forces or maintain the grasp over time.
In contrast, our approach models the full two-way coupling between an actuated hand and the object. We reconstruct not only a contact-consistent hand pose but also a control target whose induced joint torques realize those contacts and keep the grasp stable under forward simulation.

\paragraph{Physics-aware Grasping with Two-way Coupling}
A visually or geometrically plausible hand pose is not a stable grasp unless it is paired with an appropriate control strategy that holds the hand and object in place over time. As seen in a line of works in robotics~\citep{springgrasp,contactimplicitmpc,unidexgrasp,crossdex,dextrah}, stable grasping is more fundamentally a control problem, not only a geometric optimization or pose estimation problem.

Several works incorporate hand control yet do not rigorously model the two-way coupling. \citet{zhao2022stability} optimize a monocular reconstruction under a simulated stability criterion, holding the hand at the reconstructed pose with heuristic control that is never optimized, limiting the stability of the result. HOIC~\citep{hoic} imitates hand-object interaction in simulation, but it allows extra compensation force and torque applied directly to the object, so the object is not driven by two-way coupling physics alone. In contrast, D-Grasp~\citep{dgrasp} and Grasp-and-Lift~\citep{graspandlift} model strict two-way coupling: D-Grasp learns a policy that controls a hand through PD targets, and Grasp-and-Lift refines vision-aligned hand-object trajectories with physics-in-the-loop optimization. 
However, both focus on \textbf{dynamic} settings in which control can vary throughout time to maintain the grasp. Our method operates in a \textbf{static} setting: the unobserved control is represented by \textbf{one fixed PD target}. This imposes a distinct constraint on grasp reconstruction, since stability must be explained without freely changing the visible grasp or adapting the control target over a dynamic trajectory.

\section{Methodology}
\label{sec:methodology}

Our goal is to reconstruct, from a single RGB image of a human hand holding an object in place, (a) the geometry of the object, (b) the shape and pose of the hand, and (c) a representation of the joint torques and contact forces that stabilizes the hand-object system in simulation.

To be considered successful, the reconstruction must be visually consistent with the hand and object observed in the image; but in addition the grasp must be \emph{stable}: the reconstructed hand and joint torques must hold the object in place under frictional contact without noticeable motion of \textbf{either the hand or object}, as verified through physical simulation.

\paragraph{Representing the Hand and the Object} We use MANO~\citep{mano} as our visual hand model. A MANO hand mesh $\mathcal{M}_h^v = \mathcal{M}_{\mathrm{MANO}}(\mathbf{q}_h^v, \boldsymbol{\beta}_h)$ is described by both hand shape parameters (e.g. finger length)  $\boldsymbol{\beta}_h$ and hand pose $\mathbf{q}_h^v = (\mathbf{t}_h^v, \boldsymbol{\theta}_h^v) \in \mathbb{R}^{51}$, where $\mathbf{t}_h^v \in \mathbb{R}^3$ denotes the global translation of the hand and $\boldsymbol{\theta}_h^v \in \mathbb{R}^{48}$ consists of one 3D wrist rotation and fifteen 3D finger-joint rotations (three joints on each of the five fingers). For the object, we aim to reconstruct a watertight manifold triangle mesh $\mathcal{M}_o$ representing its boundary.

The above representations are purely visual and geometric. They are not equipped with physical properties and are thus required to be made simulation-ready. For the hand, we use a proxy $\mathcal{H}(\mathbf{q}_h^v; \boldsymbol{\beta}_h)$: this proxy is an articulated rigid body parameterized by the same hand pose and shape parameters as the MANO hand. For the object, we assume it is a single rigid body with known uniform density $\rho_o$.

\paragraph{What is a Stable Grasp?}
A pair of simulation-ready hand and object alone is insufficient to evaluate whether a grasp is stable. The hand pose determines the points of contact with the object, but the contact forces at those points depend on the hand joint torques (i.e., how the hand is gripping the object) which are imperceptible in the input image. Reconstructing a stable grasp thus requires inferring these unknown joint torques as well.

Given the object's acceleration due to gravity $\mathbf{g} \in \mathbb{R}^3$ (which we assume is known), conceptually, a grasp is stable if the contact forces and torques exactly balance gravity (for the object) and the joint torques (for the hand), so that neither moves over time. There are however several complications when formalizing this conceptual notion of grasp stability:

First, simulating the grasp requires a choice of contact model, time integrator, and other simulation parameters. We choose the state-of-the-art MuJoCo~\citep{MuJoCo} library as our simulation engine. The full hand-object configuration at time step $t$ is $\mathbf{x}_t = (\mathbf{q}_{h,t}, \mathbf{q}_{o,t})$, where $\mathbf{q}_{o,t} \in \mathbb{R}^6$ represents the translation and rotation of the object. Starting from the initial configuration $\mathbf{x}_0 = (\mathbf{q}_h^v, \mathbf{0})$ and zero initial configurational velocity, we evolve the system over time using MuJoCo for a horizon of $T$ time steps, and declare that the grasp is stable if $\mathbf{x}_t \approx \mathbf{x}_0$ for $t \leq T$.

Second, even if a set of joint torques and their induced contact forces exactly balance gravity during the first simulation time step, applying those same torques at every time step throughout the time horizon $T$ is not guaranteed to yield a stationary simulated trajectory $\mathbf{x}_t = \mathbf{x}_0$. Any slight force or torque residual (for example, due to floating-point or time-integration error) will start to accelerate the hand and object, so that $\mathbf{x}_t$ deviates increasingly from $\mathbf{x}_0$ over time. 
Following standard practice in robotics~\citep{andrychowicz2020learning}, instead of directly reconstructing fixed joint torques, we reconstruct a \emph{target hand pose} $\mathbf{q}_h^{\mathrm{target}} \in \mathbb{R}^{51}$. The joint torques are given by a PD (proportional-derivative) controller at each simulation timestep. Intuitively, it is a separate hand pose (typically intersecting the object) that the hand attempts to reach via joint actuation.

\begin{figure}[!htbp]
    \centering
    \includegraphics[width=\linewidth]{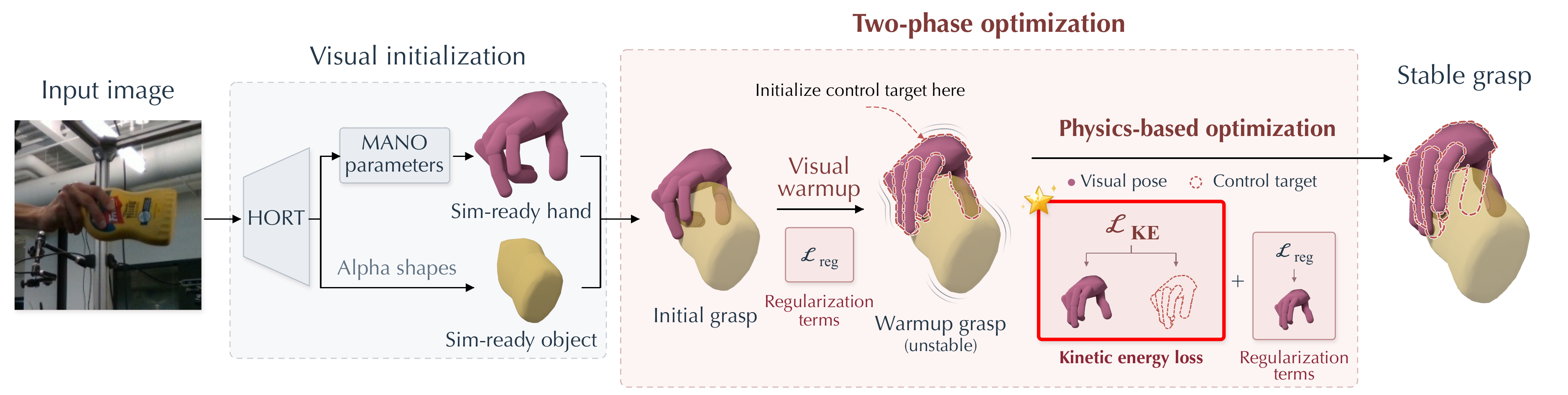}
    \caption{Pipeline overview. We first obtain a visual reconstruction of the hand and the object and make them simulation-ready. Then we refine the grasp with two-phase optimization. Starting from the second phase, a physics-based kinetic energy (KE) loss jointly optimizes the visual hand pose and the control target. The visual hand pose is also constrained by regularization terms.}
    \label{fig:pipeline}
\end{figure}

\subsection{Pipeline}
\label{sec:pipeline}

Given a single RGB image $I$ together with object density $\rho_o$ and gravity $\mathbf{g}$, our pipeline reconstructs: a mesh $\mathcal{M}_o$ of the object; MANO hand shape parameters $\boldsymbol{\beta}_h$; and visual hand pose parameters $\mathbf{q}_h^v$, all of which are consistent with the image; and a target hand pose $\mathbf{q}_h^{\mathrm{target}}$ such that the object and hand remain stationary after $T$ time steps of physical simulation.

As shown in Fig.~\ref{fig:pipeline}, our pipeline consists of two steps. 
\emph{Visual initialization} estimates $\mathcal{M}_o$, $\boldsymbol{\beta}_h$, and $\mathbf{q}_h^v$ from the image without considering physics. \emph{Differentiable physics-based optimization} (described below in Section~\ref{sec:physics-opt}) refines the estimated pose. Here we freeze $\mathcal{M}_o$ and $\boldsymbol{\beta}_h$ and seek a target pose $\mathbf{q}_h^{\mathrm{target}}$ that explains grasp stability. The visual pose $\mathbf{q}_h^v$ is jointly optimized at the same time.

We use HORT~\citep{HORT} to perform the initial hand and object reconstruction. The hand branch of HORT predicts MANO hand pose $\mathbf{q}_h^v$ and shape parameters $\boldsymbol{\beta}_h$ from the input image; its object branch predicts a 3D object point cloud. We then reconstruct a closed object surface from the point cloud using alpha-shape meshing~\citep{Edelsbrunner94}, and convexly decompose the mesh using CoACD~\citep{Wei2022} to make it simulation-ready in MuJoCo.

\begin{figure}[!htbp]
    \centering
    \includegraphics[width=\linewidth]{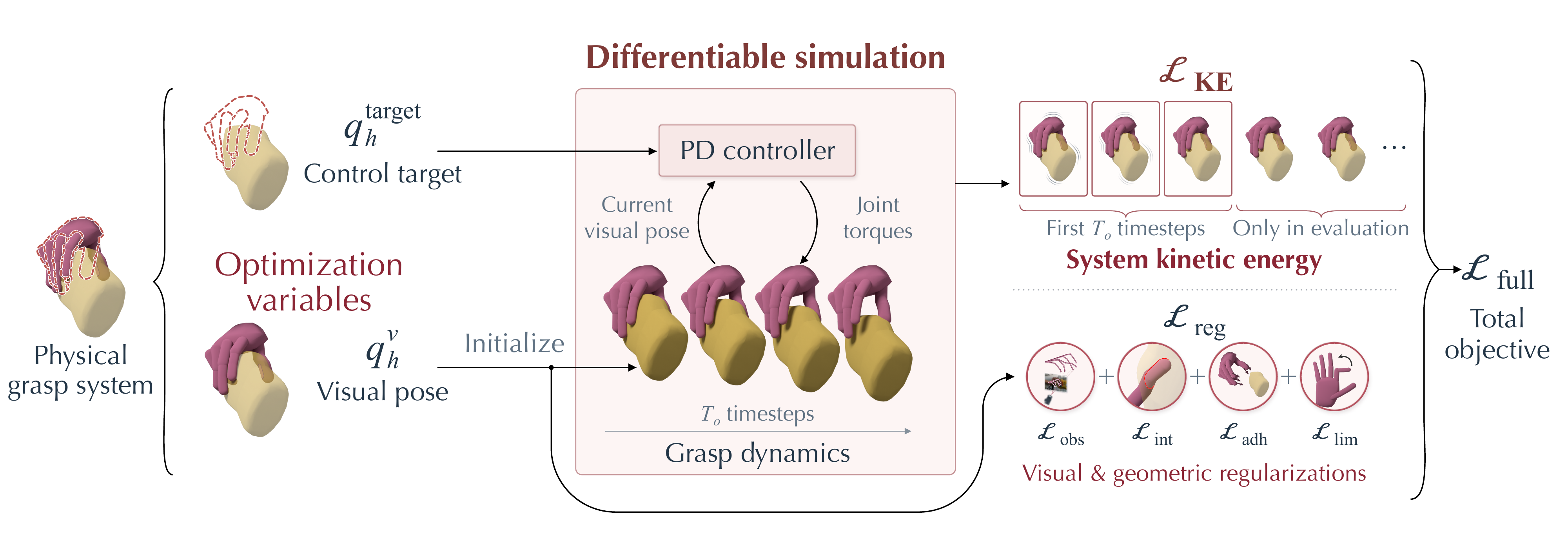}
    \caption{Differentiable physics-based optimization. The visual hand pose $\mathbf{q}_h^v$ initializes the hand--object system, while the fixed control target $\mathbf{q}_h^{\mathrm{target}}$ drives a PD controller during simulation. During physics-based refinement, both poses are jointly optimized using the total hand--object kinetic energy accumulated over the first $T_o$ timesteps and visual and geometric regularization of the visual pose.}
    \label{fig:optimization}
\end{figure}

\subsection{Differentiable Physics-based Optimization}
\label{sec:physics-opt}
As shown in Fig.~\ref{fig:optimization}, the second step of our pipeline jointly refines the visual hand pose $\mathbf{q}_h^v$ and computes the target hand pose $\mathbf{q}_h^{\mathrm{target}}$.
We do so by optimizing an objective function that directly measures stability, along with visual and geometric regularizers that keep the hand pose consistent with the image and free of penetrations into the object volume.

\paragraph{Kinetic Energy Objective}
Recall that a visual pose and target pose pair $\left(\mathbf{q}_h^v, \mathbf{q}_h^{\mathrm{target}}\right)$ is stable if the generalized configuration changes negligibly between time steps: $\mathbf{x}_{t+1} \approx \mathbf{x}_t \approx \mathbf{x}_0.$
Since the natural, coordinate-free metric on configurational displacements is kinetic energy, we use the total kinetic energy of the simulated hand-object system, integrated over $T_o$ steps of forward simulation (where $T_o \ll T$), to quantify stability. We use a much shorter horizon because instability emerges within the first few timesteps. This also makes efficient backpropagation possible in practice.

Let $\dot{\mathbf{x}}_t$ denote the system's configurational velocity, and $\mathbf{M}(\mathbf{x}_t)$ denote the generalized mass matrix (which includes both mass and inertia tensors for the hand and the object). We then accumulate kinetic energy over the first $T_o$ simulation steps and define the physical stability term as
\[
\mathcal{L}_{\mathrm{KE}} = \sum_{t=1}^{T_o} \frac{1}{2}\dot{\mathbf{x}}_t^{\top} \mathbf{M}(\mathbf{x}_t) \dot{\mathbf{x}}_t.
\]

\paragraph{Visual and Geometric Objectives}
We regularize the visual hand pose $\mathbf{q}_h^v$ with four terms that keep it
consistent with the image and free of penetration into the object. A 2D
reprojection term $\mathcal{L}_{\mathrm{obs}}$ penalizes the $L_1$ distance between the
projected 3D hand joints and the 2D joint locations estimated by HORT. With $N$ points sampled on the hand proxy $\mathcal{H}$ and computing their signed distances to
the object surface, an interpenetration term $\mathcal{L}_{\mathrm{int}}$
penalizes points lying inside the object, while an adhesive term
$\mathcal{L}_{\mathrm{adh}}$ (following prior visual grasp optimization methods like~\citep{easyhoi}) pulls exterior
non-palm points toward the surface to encourage contact formation. Palm points
are excluded because the rigid proxy palm does not contribute to stability the way a
deformable human palm does. A joint-limit term $\mathcal{L}_{\mathrm{lim}}$
discourages implausible configurations while leaving feasible poses unaffected.
We combine the four terms with associated weights into the regularization loss
$\mathcal{L}_{\mathrm{reg}}$.

\paragraph{Two-phase Optimization}
We use the visual hand pose provided by HORT in the Visual Initialization step as the initial guess for $\mathbf{q}_h^v$. We then optimize in two phases. In the first phase, we perform a visual warm-up in which only $\mathcal{L}_{\mathrm{reg}}$ is minimized. This stage reduces severe penetrations and establishes a visual hand pose with plausible hand-object contact before simulation-based refinement. During this phase, $\mathbf{q}_h^v$ is updated to $\mathbf{q}_h^{\mathrm{warmup}}$. In the second phase, we initialize the target pose $\mathbf{q}_h^{\mathrm{target}}$ to be $\mathbf{q}_h^{\mathrm{warmup}}$, activate the kinetic-energy term, and optimize the full objective
\[
\mathcal{L}_{\mathrm{full}}
=
\lambda_{\mathrm{KE}}
\frac{\mathcal{L}_{\mathrm{KE}}(\mathbf{q}_h^v,\mathbf{q}_h^{\mathrm{target}})}{\mathcal{L}_{\mathrm{KE}}^{(0)}}
+
\lambda_{\mathrm{reg}}
\frac{\mathcal{L}_{\mathrm{reg}}(\mathbf{q}_h^v)}{\mathcal{L}_{\mathrm{reg}}^{(0)}}.
\]
Here, $\mathcal{L}_{\mathrm{KE}}^{(0)}$ and $\mathcal{L}_{\mathrm{reg}}^{(0)}$ are the values of the two loss bundles evaluated once at the beginning of the second phase, i.e., from the warm-up result. This normalization compensates for any scale mismatch between the two loss terms, so that their relative influence is controlled by $\lambda_{\mathrm{reg}}$ and $\lambda_{\mathrm{KE}}$ rather than by raw loss magnitude alone.

\FloatBarrier
\section{Experiments}
\label{sec:experiments}

\subsection{Experimental Setup}

\paragraph{Datasets}
We evaluate on DexYCB~\citep{dexycb}, HO3D~\citep{ho3d}, and
CORe50~\citep{core50}, using the HORT~\citep{HORT} test splits for
DexYCB and HO3D. Our benchmark contains $242$ images from $81$
DexYCB sequences, $100$ frames from $10$ HO3D sequences, and $20$
manually selected non-laboratory CORe50 images with diverse objects
and hand poses. DexYCB images are final frames from selected camera
views with clear hand--object visibility. HO3D frames are temporally
subsampled near sequence ends and can depict ongoing manipulation.

Object densities follow YCB statistics~\citep{ycbobjects} for DexYCB
and HO3D. For CORe50, which provides neither 3D annotations nor object
masses, we use $1000\,\mathrm{kg/m^3}$. Gravity is approximated by the
camera-down direction with magnitude $9.8\,\mathrm{m/s^2}$.

\paragraph{Implementation Details}
We use MuJoCo XLA (MJX)~\citep{MuJoCo,JAX} and automatically generate
each hand proxy from the HORT-estimated MANO shape using a fixed
template, with a box palm~\citep{dgrasp} and capsule finger
links~\citep{keepitsmpl}. Each refinement iteration differentiates through
$T_o=20$ simulation steps at $2$ ms. Physics-based optimization early-stops after
200 iterations without sufficient loss improvement. Evaluation uses $0.2$ ms steps over $T=5$ s to reduce integration
errors, with unified PD gains.

\subsection{Metrics}

\paragraph{Physical Stability}
We measure hand motion from initial to final states by 3D MPJPE~\citep{ionescu2013human3} between the initial and final joint
positions of the simulation, together with net object center-of-mass translation and
rotation angle. Success Rate is the fraction of grasps whose object
center remains within $1$ cm of its initial position throughout
simulation. Drop Time is the mean time to first exceed this threshold
among failed grasps.

\paragraph{Visual Stability}
We compare initial and final image projections using object-mask IoU
and hand PCK~\citep{yang2012articulated}. PCK@$\tau$ is the fraction
of the $21$ projected joints displaced by less than $\tau$ in
normalized image coordinates.

\begin{table}[!t]
\centering
\begingroup
\setlength{\tabcolsep}{1pt}
\renewcommand{\arraystretch}{1.0}
\newlength{\qualcellside}
\newcommand{\qualimage}[1]{\raisebox{-.5\height}{%
  \includegraphics[width=\qualcellside,height=\qualcellside,keepaspectratio]{#1}}}
\newcommand{\qualhead}[1]{\parbox[c]{\qualcellside}{\centering\fontsize{6.5}{7.4}\selectfont #1}}
\newcommand{\qualrow}[4]{%
  #1 & \qualimage{figures/figure1/#2/#3/input.png}
     & \qualimage{figures/figure1/#2/#3/t0p00.png}
     & \qualimage{figures/figure1/#2/#3/t5p00.png}
     & \qualimage{figures/figure1/target_#4.png}\\[2pt]}
\newcommand{\qualexample}[3]{%
  \begin{minipage}[t]{0.49\linewidth}
  \centering\fontsize{7}{8}\selectfont
  \setlength{\qualcellside}{0.205\linewidth}
  \textbf{#1}\par\vspace{3pt}
  \begin{tabular}{@{}l@{\hspace{2pt}}cccc@{}}
    & \qualhead{Input} & \qualhead{Reconstruction}
    & \qualhead{Final} & \qualhead{Target}\\[2pt]
    \qualrow{PPO}{#2}{dgrasp}{#3d}
    \qualrow{Pinching}{#2}{pinching}{#3p}
    \qualrow{\textbf{Ours}}{#2}{ours}{#3o}
  \end{tabular}
  \end{minipage}}
\newcommand{\qualplaceholder}{\raisebox{\dimexpr.5\depth-.5\height\relax}{%
  \begingroup\setlength{\fboxsep}{0pt}%
  \colorbox{black!5}{\parbox[c][\qualcellside][c]{\qualcellside}{\centering\textcolor{black!30}{---}}}%
  \endgroup}}
\qualexample{(a) DexYCB: box}{dexycb/1}{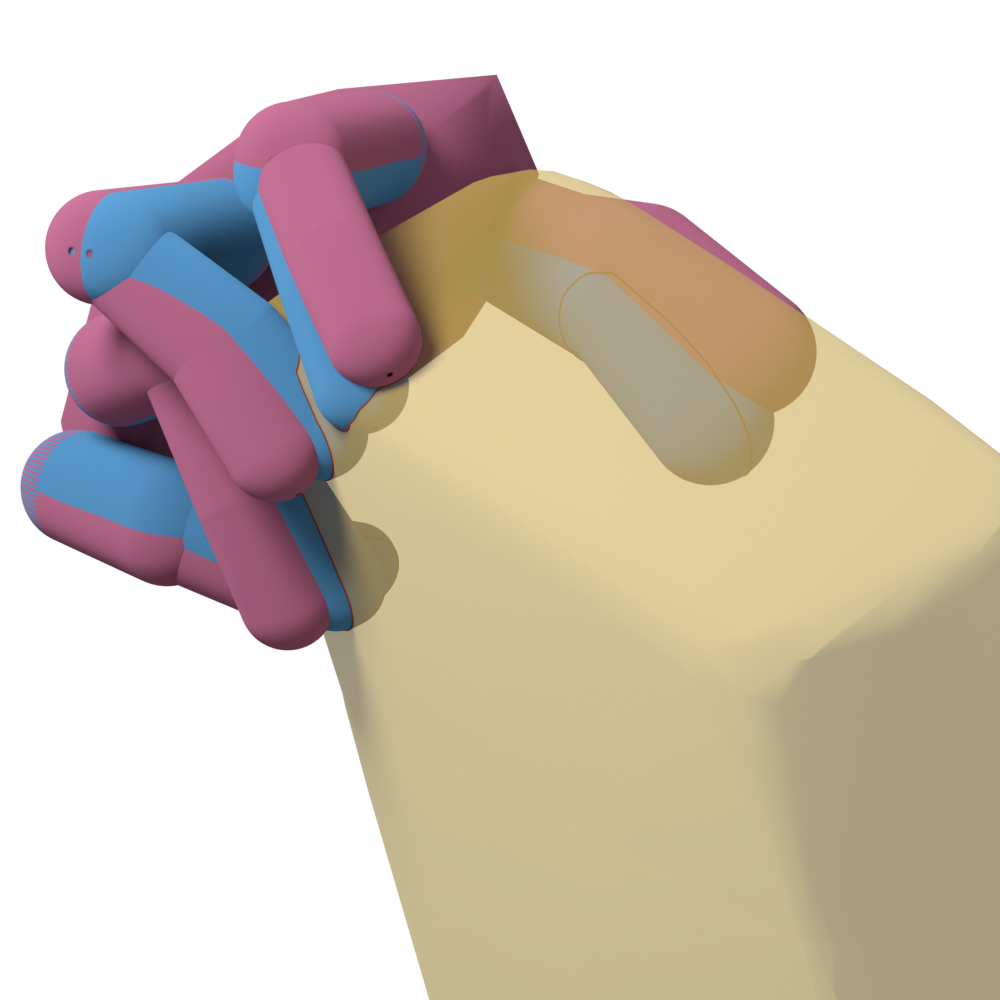}\hfill
\qualexample{(b) DexYCB: drill}{dexycb/2}{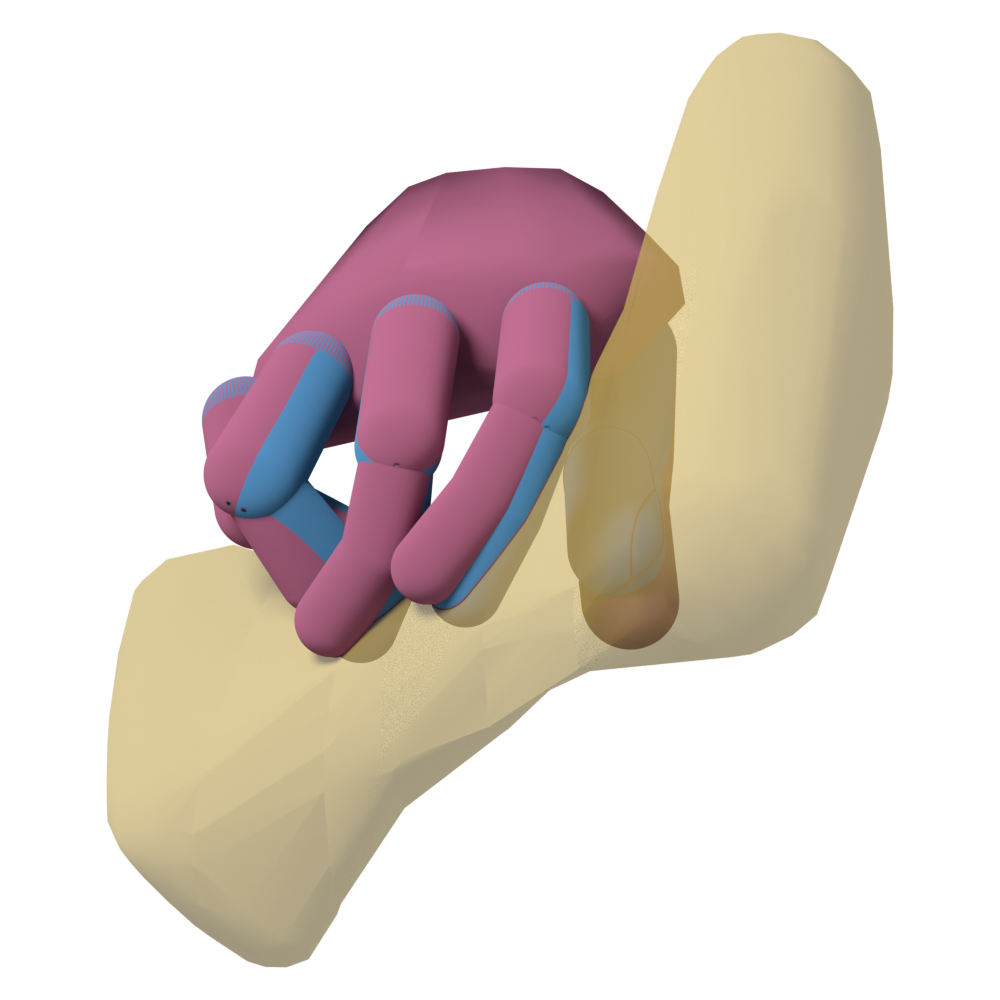}
\par\vspace{6pt}
\qualexample{(c) HO3D}{ho3d/4}{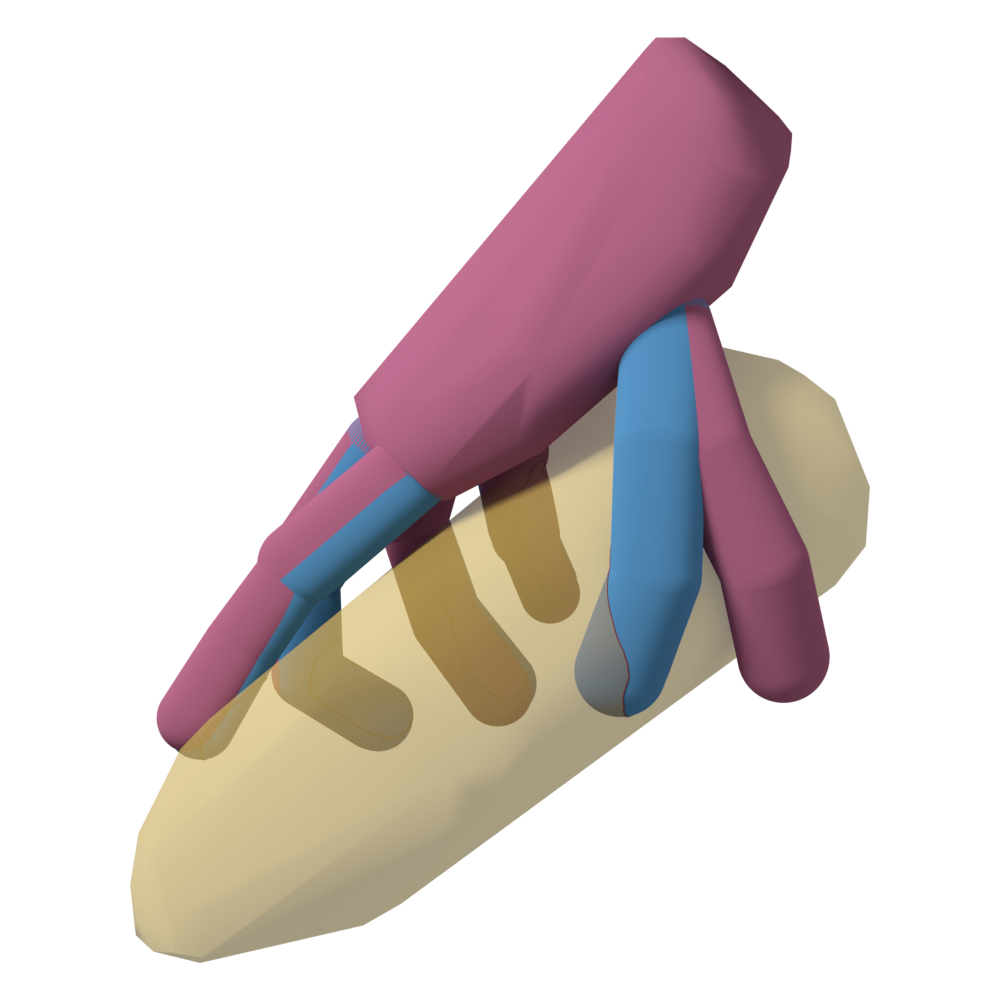}\hfill
\begin{minipage}[t]{0.49\linewidth}
  \centering\fontsize{7}{8}\selectfont
  \setlength{\qualcellside}{0.205\linewidth}
  \textbf{(d) CORe50: mug}\par\vspace{3pt}
  \begin{tabular}{@{}l@{\hspace{2pt}}cccc@{}}
    & \qualhead{Input} & \qualhead{Reconstruction}
    & \qualhead{Final} & \qualhead{Target}\\[2pt]
    PPO & \qualimage{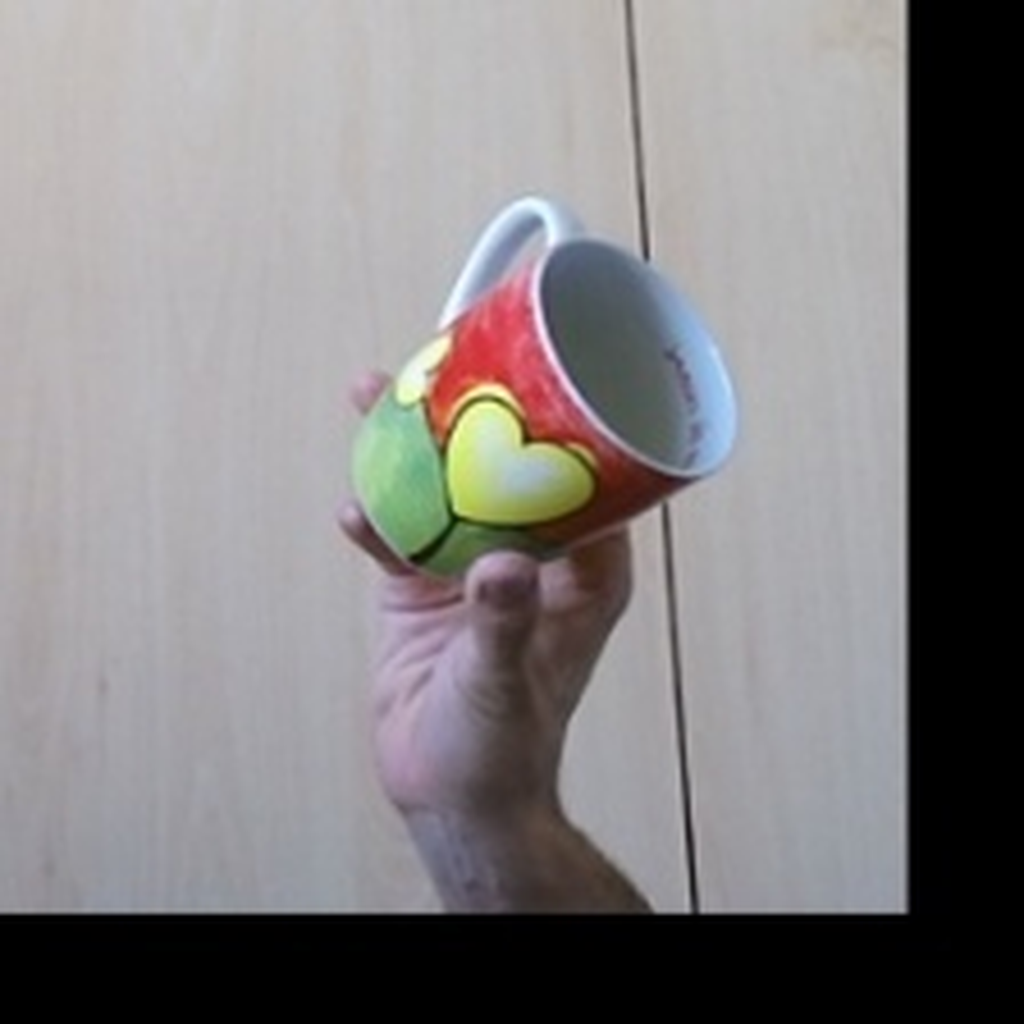} & \qualimage{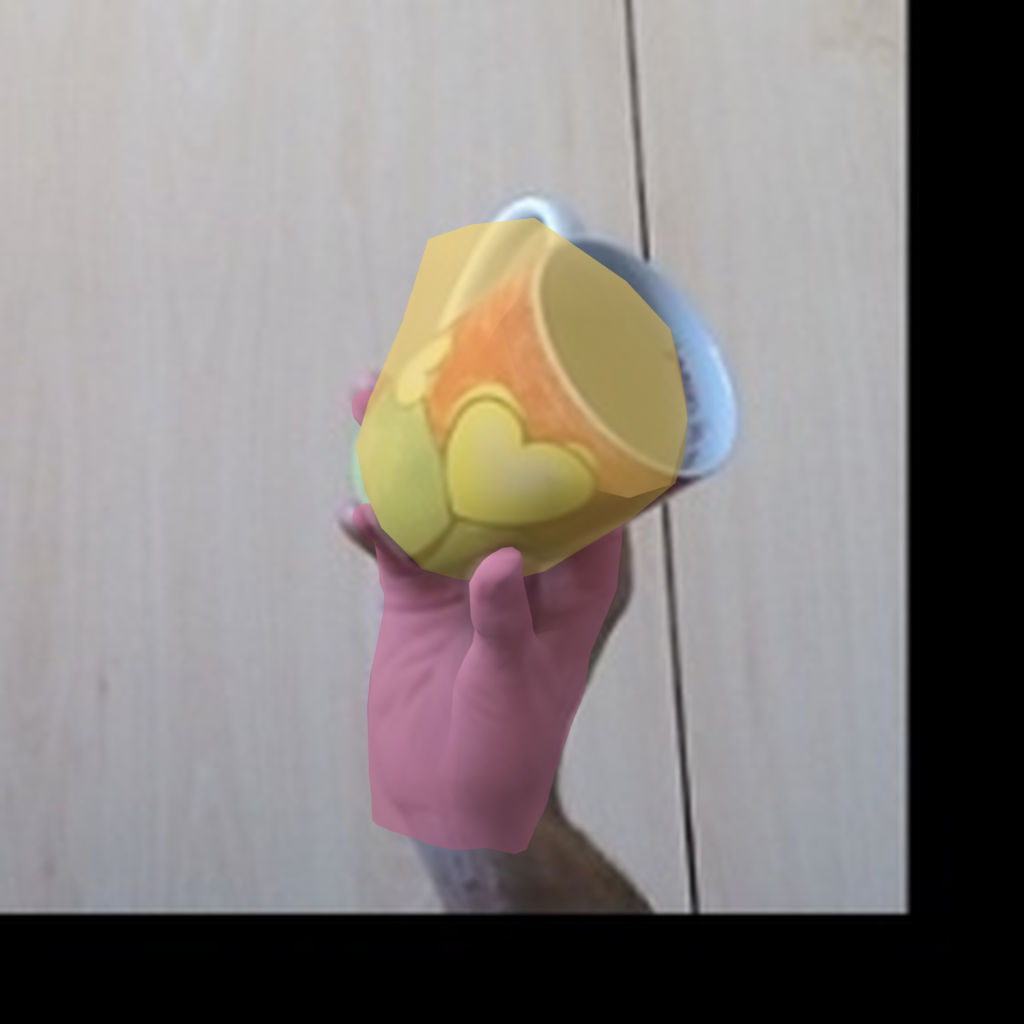} & \qualimage{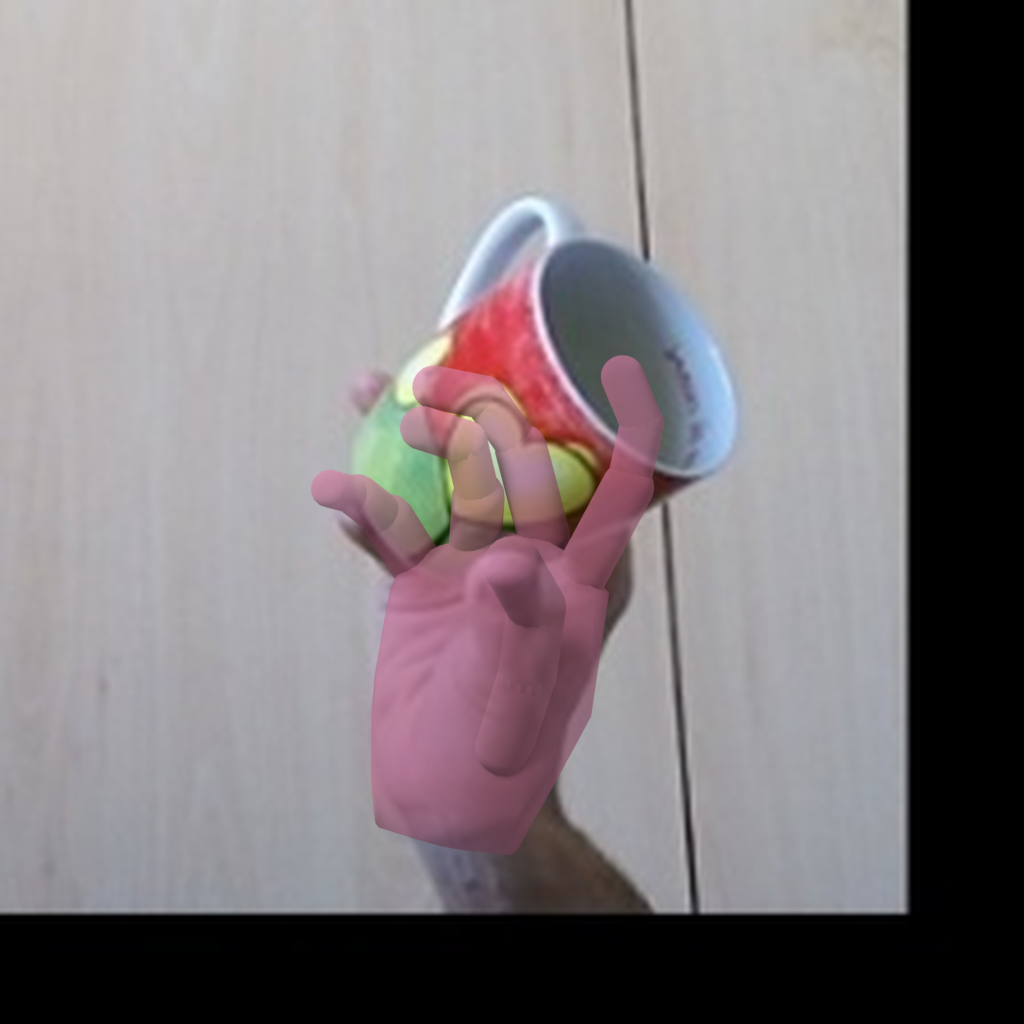} & \qualimage{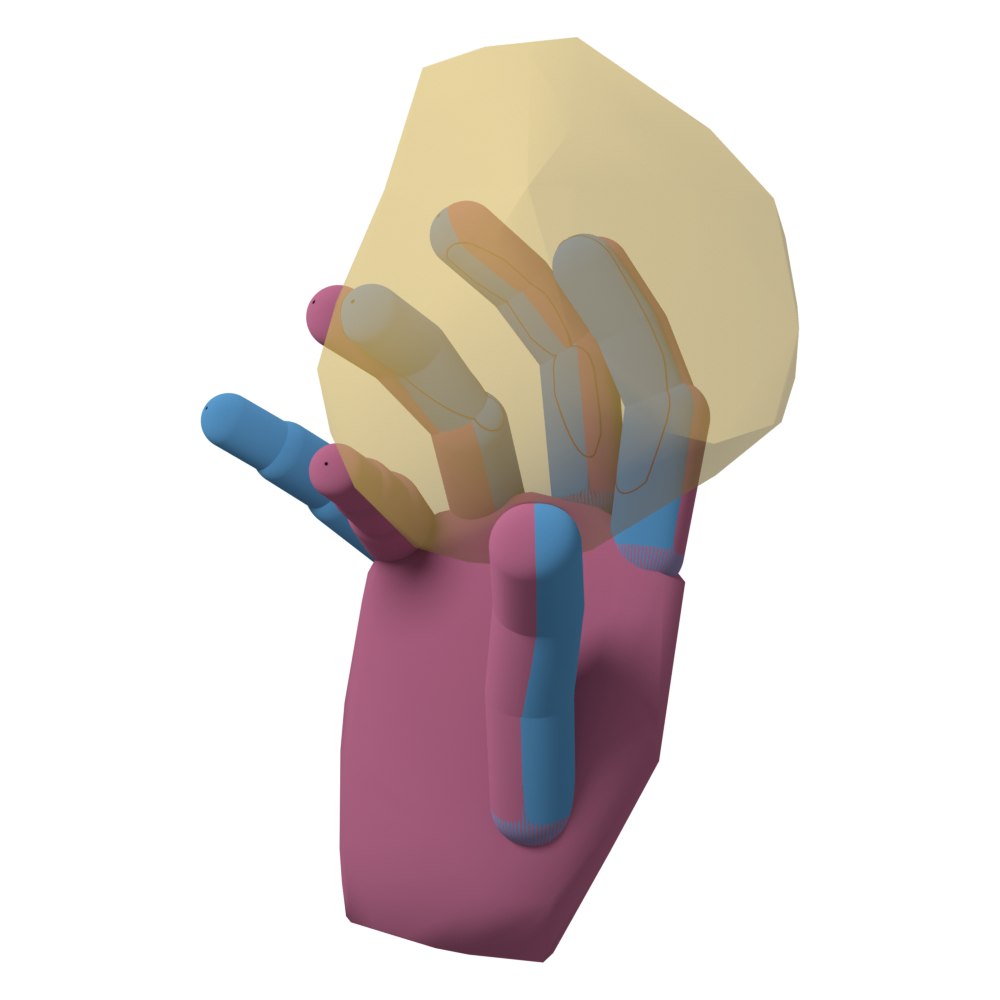}\\[2pt]
    Pinching & \qualimage{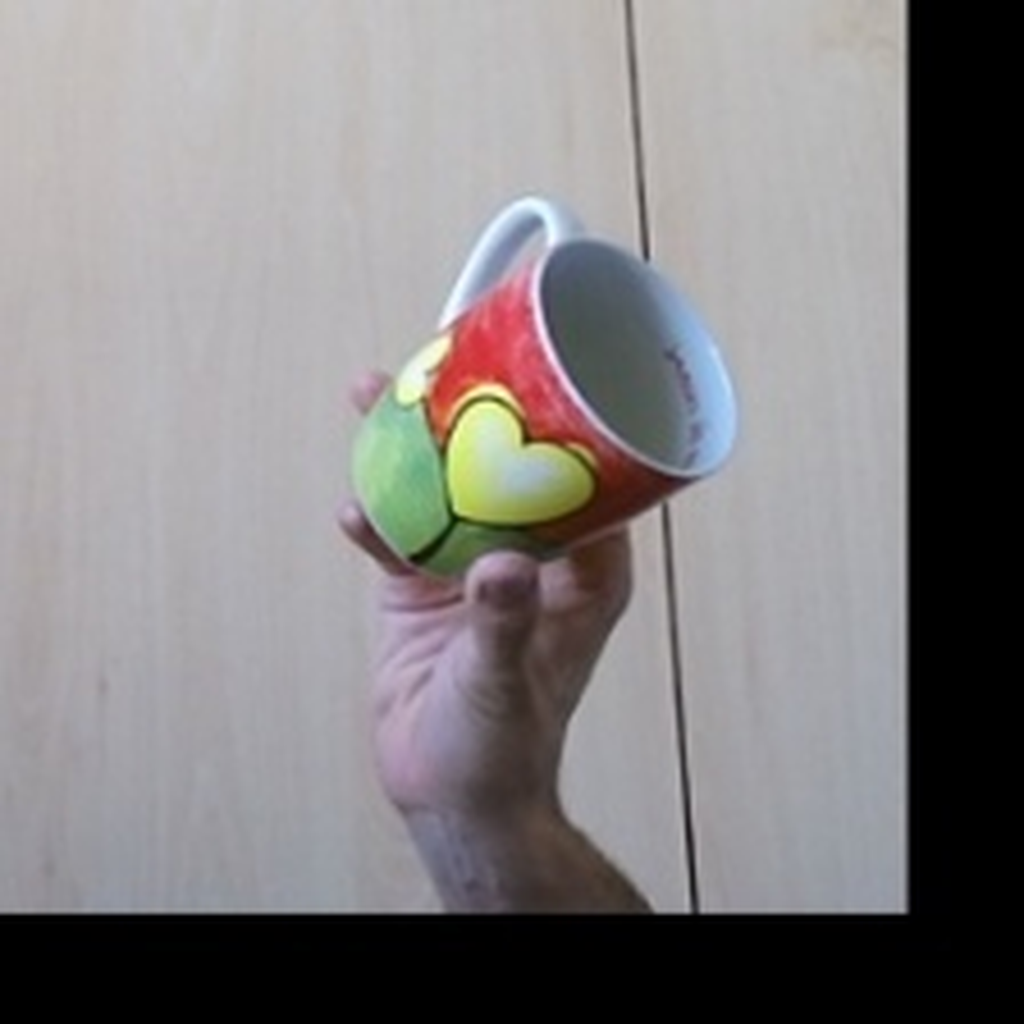} & \qualimage{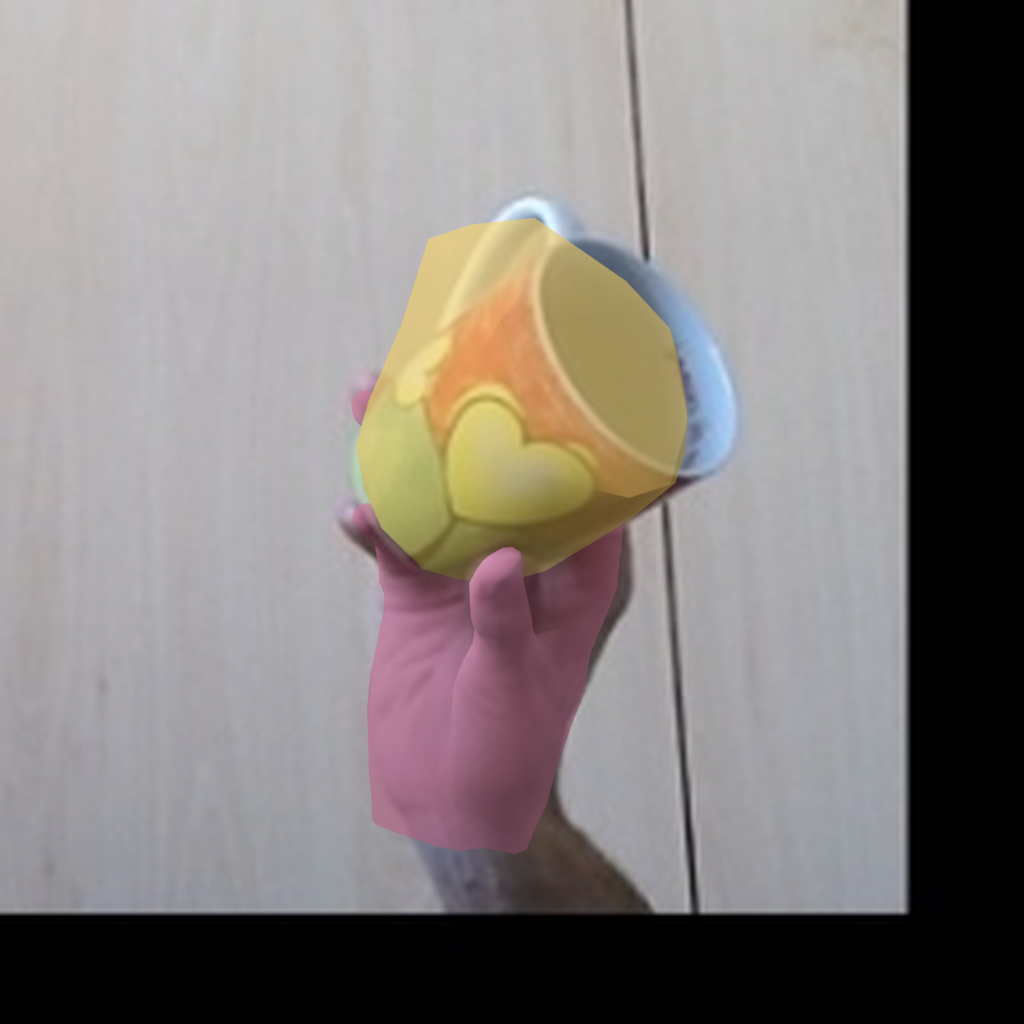} & \qualimage{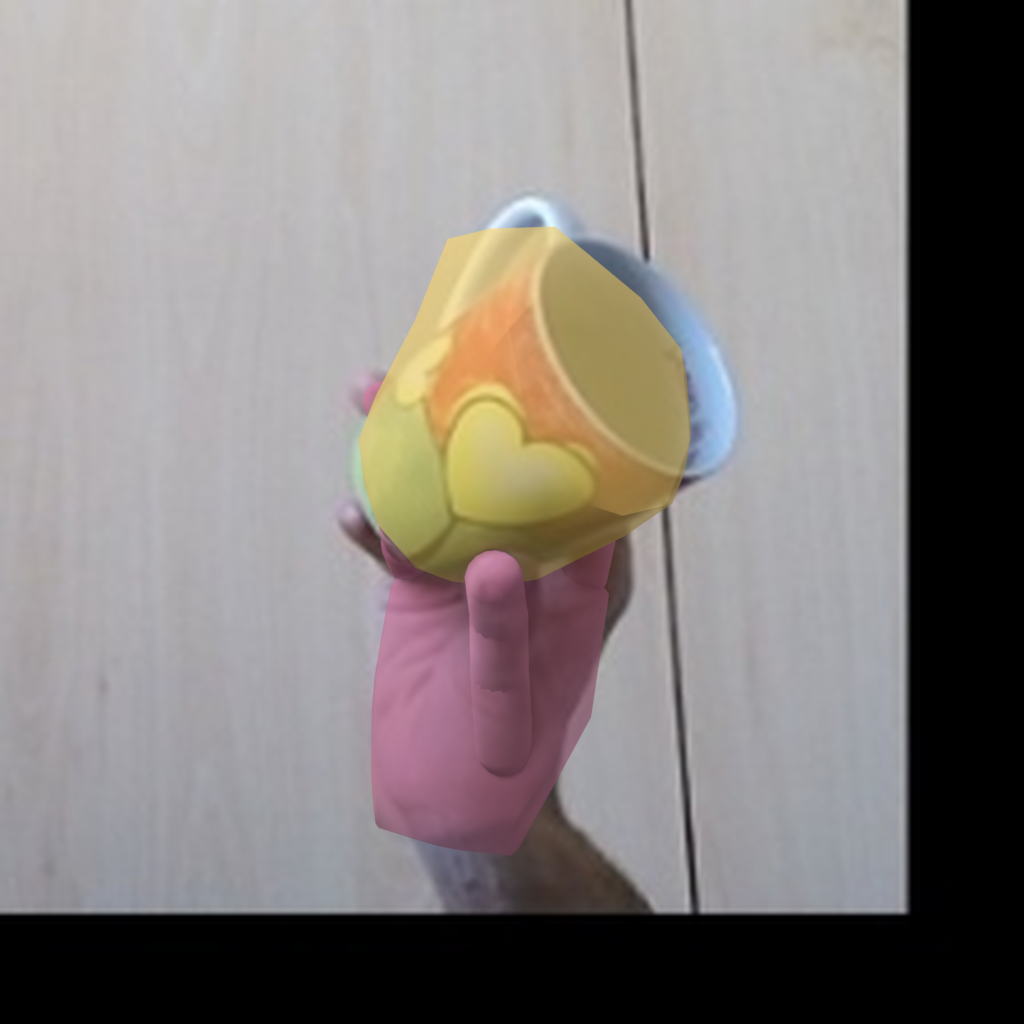} & \qualimage{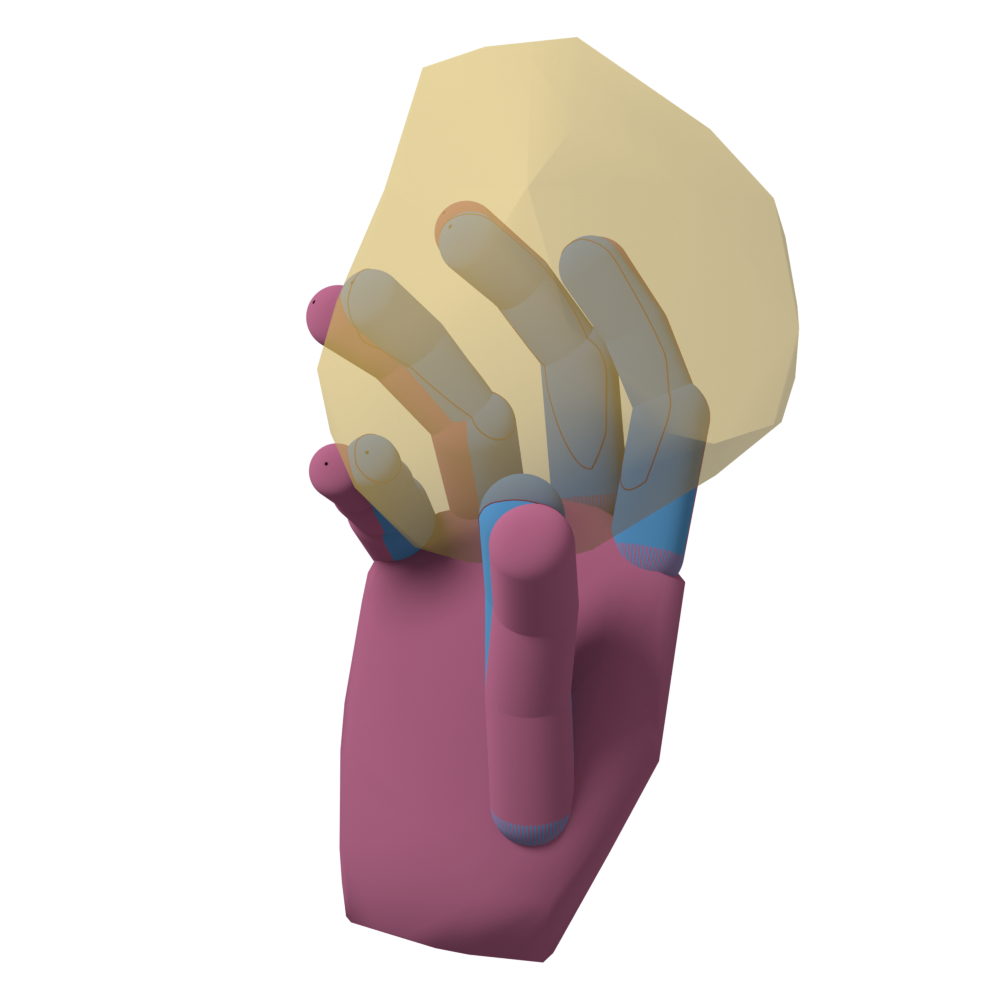}\\[2pt]
    \textbf{Ours} & \qualimage{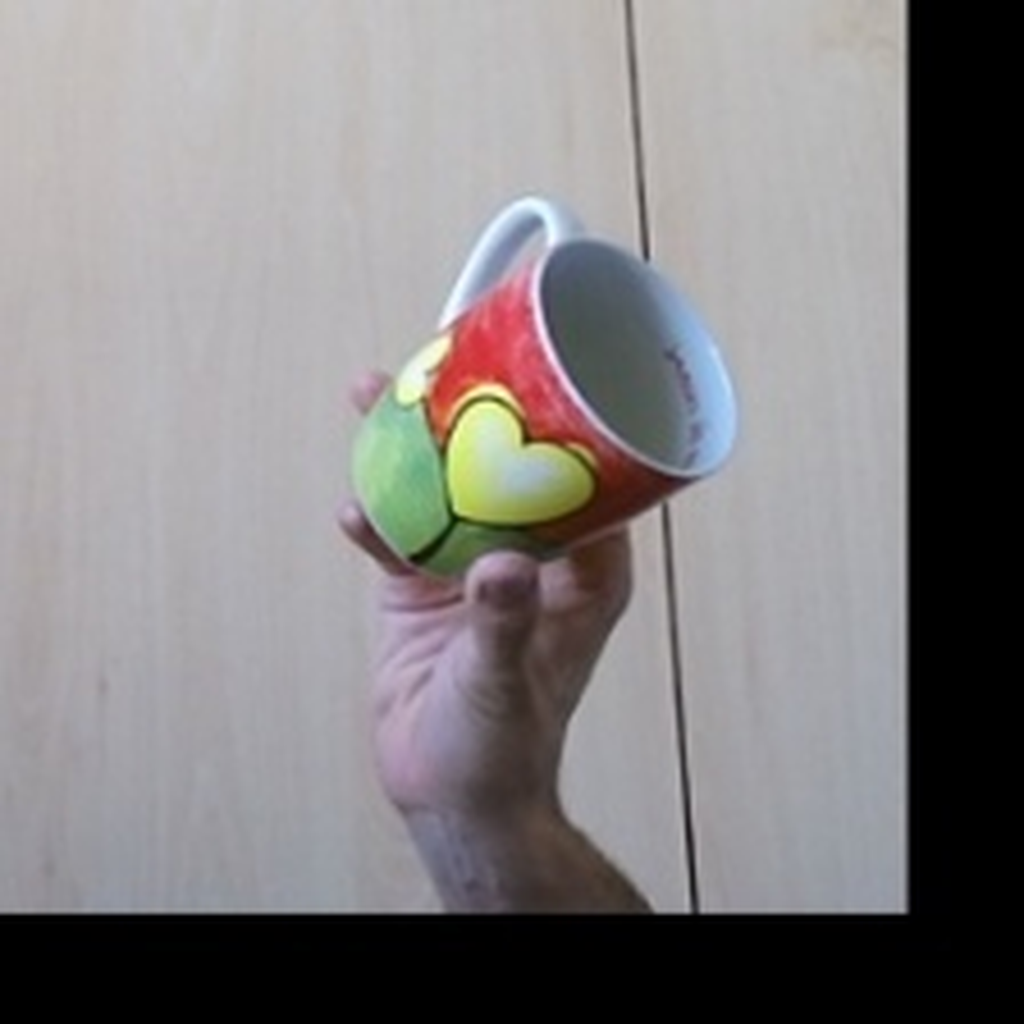} & \qualimage{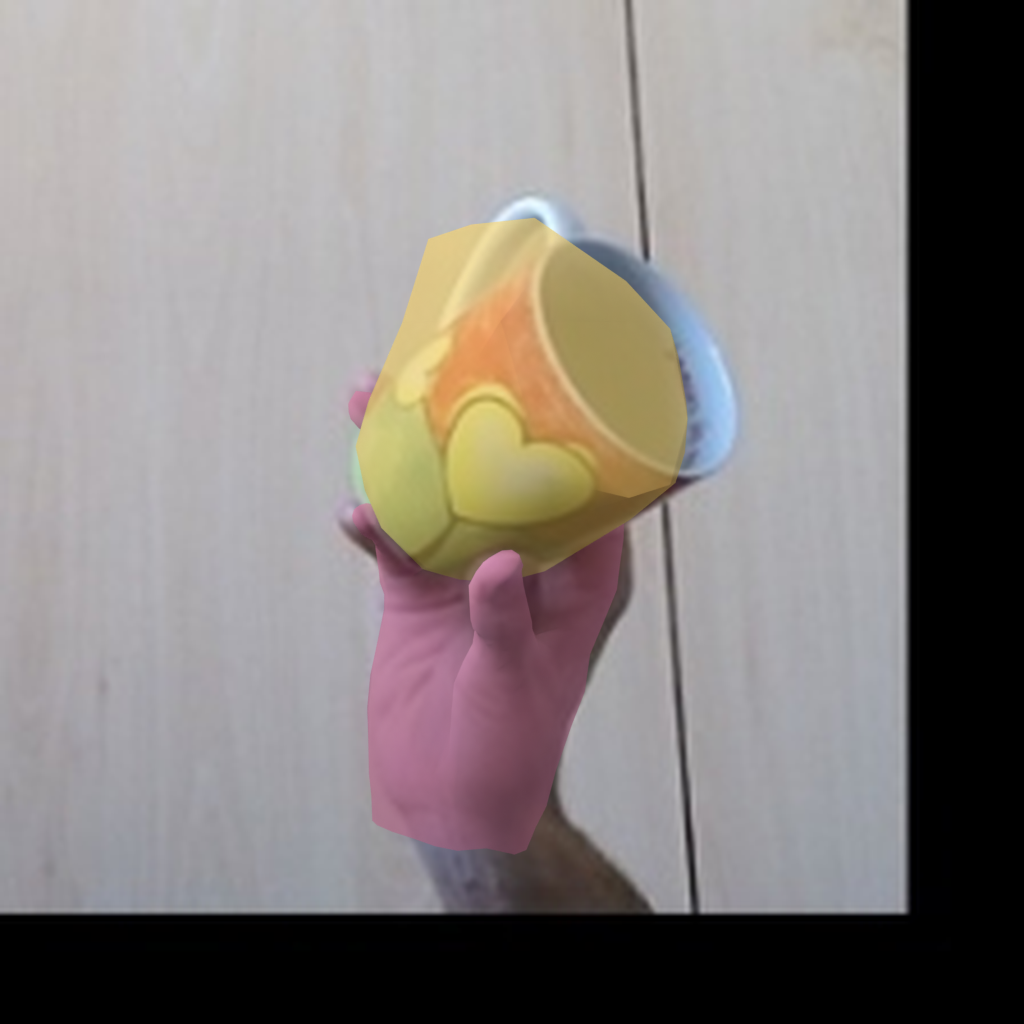} & \qualimage{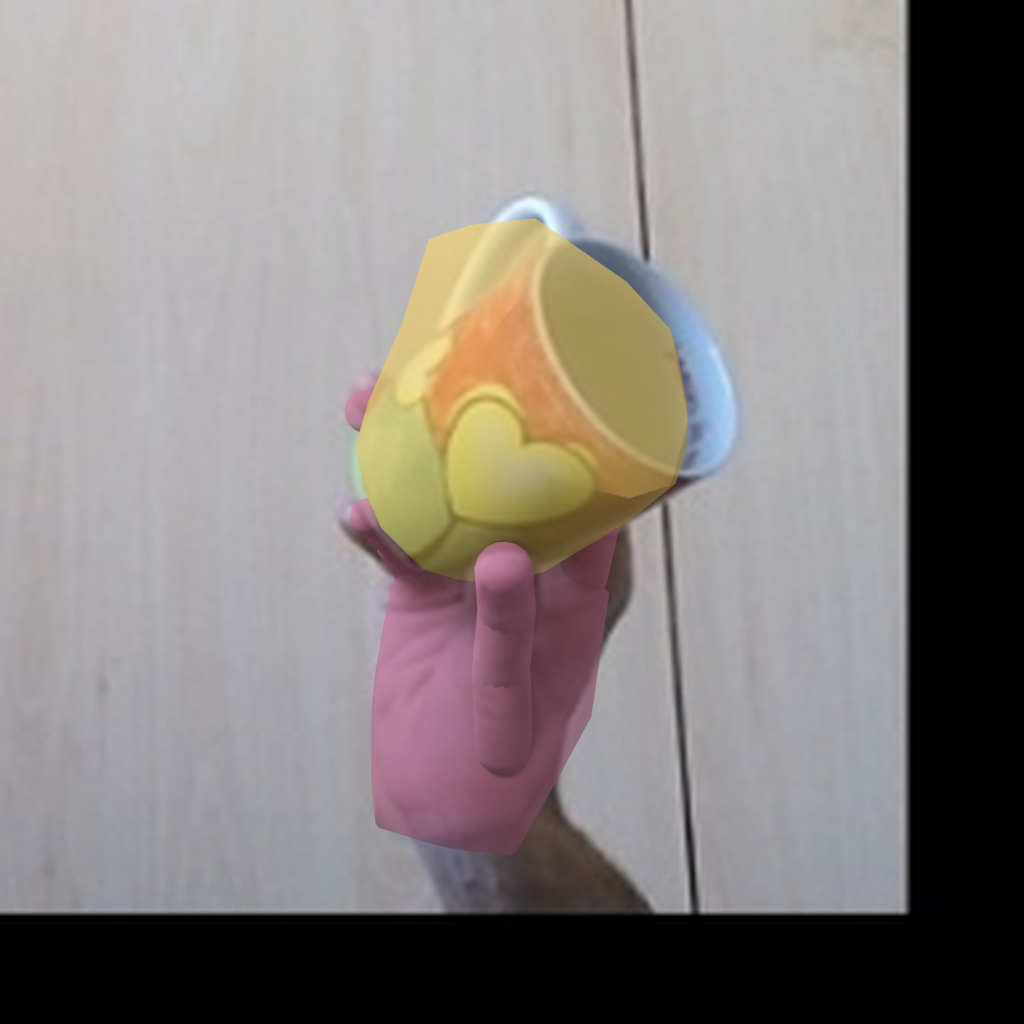} & \qualimage{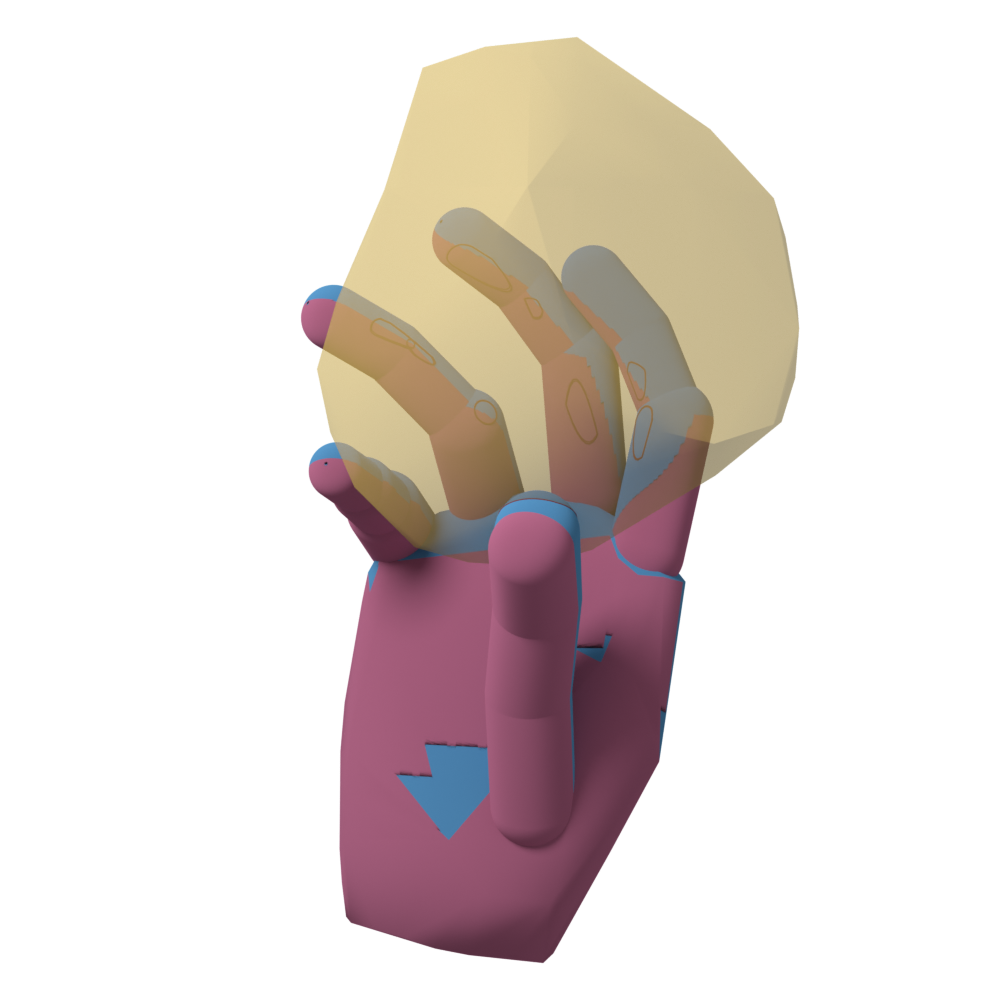}\\[2pt]
  \end{tabular}
\end{minipage}
\endgroup
\caption{Qualitative comparisons. Examples from all three datasets show object drops or visible hand--object motion with PPO and Pinching, while our method preserves stable grasps. Even when the baselines retain the object, their larger shifts are readily noticeable. The target visualizations (control targets shown in blue) show how our method flexibly adjusts the control target to achieve stability, compared with the prescribed inward displacement of Pinching.}
\label{tab:qualitative}
\end{table}

\paragraph{Reconstruction Quality}
For DexYCB and HO3D, we compare $\mathbf{q}_h^v$ with their annotations
using 3D MPJPE and 2D PCK. These MANO fits to RGB-D
observations~\citep{dexycb,ho3d} do not account for grasp dynamics
and can contain interpenetration~\citep{deepsimho,ho3dv3}.
Contact depends on fine surface geometry and tissue
deformation~\citep{contactopt}; stable contact with our rigid proxy
and reconstructed object can therefore require pose adjustments
relative to these visual references.

\subsection{Baseline Comparisons}
\label{sec:baseline}

We compare StableGrasp with alternative approaches for stabilizing
reconstructed grasps through hand control: heuristic pinching, and a learned instance-specific
PPO baseline. Each baseline specifies
how the hand is controlled, thus enabling comparison of grasp stability
under forward simulation of the coupled hand--object system. For all baselines, we initialize the hand pose to that computed by our visual warm-up optimization.

\paragraph{Heuristic \textit{Pinching}}
A commonly-used~\citep{dexgraspnet,easyhoi} heuristic for stabilizing a grasp without physics-driven optimization is to specify a target hand pose $\mathbf{q}_h^{\mathrm{target}}$ where each finger has been displaced some fixed distance into the object along the contact normal direction. A PD controller is then used to compute joint torques from the target pose. We adapt the approach of DexGraspAnything~\citep{dexgraspanything} as a representative method in this category. 

\paragraph{\textit{PPO}}
We study learned grasp stabilization with an instance-specific PPO baseline
adapted from D-Grasp~\citep{dgrasp}. Each policy starts from the warm-up grasp, is individually trained and is allowed to update finger PD targets online over time in the evaluation simulation. In our experiments, we drop out 5 examples where policy training fails with numerical errors.

\subsubsection{Qualitative Results}
Table~\ref{tab:qualitative} compares the input images, reconstructed
grasps, final simulated states, and a visualization of the optimized target pose.  Not dropping an object alone does not certify a stable grasp, because a hand that drifts noticeably is not the stable hand in the image. StableGrasp substantially
reduces motion of \emph{both the hand and object} compared with the baselines.

\subsubsection{Quantitative Results}
Table~\ref{tab:quantitative} shows that StableGrasp achieves higher success
rates and lower hand and object motion than both baselines on all three
datasets, together with better image-space stability. Note that Success Rate does not tell the full story of grasp stability: an object can significantly shift over time away from its pose as depicted in the input image without dropping completely. The metrics in Table~\ref{tab:quantitative} demonstrate that StableGrasp successfully reconstructs grasps that remain stable under rigorous forward physics simulation.

\begin{table}[!t]
\centering
\small
\setlength{\tabcolsep}{2.5pt}
\renewcommand{\arraystretch}{1.08}
\caption{Physical and visual stability (mean $\pm$ standard deviation) on 357 common examples (DexYCB: 239; HO3D: 99; CORe50: 19), excluding five PPO training failures. All methods use a 1 cm displacement threshold. Motion, IoU, and PCK compare initial and final saved states of successful grasps; Success Rate uses all examples. Our method achieves substantially higher success rates while maintaining much more stable hand and object poses.}
\label{tab:quantitative}
\begin{adjustbox}{max width=\linewidth}
\begin{tabular}{llcccccccc}
\toprule
Dataset & Method & \shortstack{MPJPE\\(mm) $\downarrow$} & \shortstack{Obj. trans.\\(mm) $\downarrow$} & \shortstack{Obj. rot.\\($^\circ$) $\downarrow$} & \shortstack{Object\\IoU $\uparrow$} & \shortstack{PCK $\uparrow$\\@0.025} & \shortstack{PCK $\uparrow$\\@0.05} & \shortstack{Success\\rate $\uparrow$} & \shortstack{Drop time\\(s) $\uparrow$} \\
\midrule
DexYCB & PPO & \statpair{4.6}{2.5} & \statpair{4.5}{2.8} & \statpair{6.40}{4.57} & \statpair{0.88}{0.09} & \statpair{0.86}{0.10} & \statpair{0.93}{0.07} & 39.33\% & \statpair{0.28}{0.70} \\
 & Pinching & \statpair{2.7}{0.8} & \statpair{5.9}{1.9} & \statpair{4.66}{2.79} & \statpair{0.87}{0.08} & \statpair{0.95}{0.06} & \statpair{\textbf{1.00}}{0.01} & 64.02\% & \statpair{0.52}{0.91} \\
 & \textbf{Ours} & \statpair{\textbf{0.2}}{0.2} & \statpair{\textbf{1.2}}{1.9} & \statpair{\textbf{1.07}}{1.57} & \statpair{\textbf{0.97}}{0.05} & \statpair{\textbf{1.00}}{0.00} & \statpair{\textbf{1.00}}{0.00} & \textbf{87.45}\% & \statpair{\textbf{0.80}}{1.09} \\
\midrule
HO3D & PPO & \statpair{4.2}{1.5} & \statpair{5.8}{2.2} & \statpair{9.87}{5.18} & \statpair{0.84}{0.05} & \statpair{0.81}{0.10} & \statpair{0.92}{0.08} & 32.32\% & \statpair{\textbf{0.59}}{0.94} \\
 & Pinching & \statpair{2.6}{0.6} & \statpair{5.3}{1.8} & \statpair{3.40}{2.47} & \statpair{0.89}{0.05} & \statpair{0.93}{0.07} & \statpair{\textbf{1.00}}{0.00} & 63.64\% & \statpair{0.53}{1.06} \\
 & \textbf{Ours} & \statpair{\textbf{0.2}}{0.2} & \statpair{\textbf{1.2}}{1.4} & \statpair{\textbf{0.88}}{1.03} & \statpair{\textbf{0.97}}{0.03} & \statpair{\textbf{1.00}}{0.00} & \statpair{\textbf{1.00}}{0.00} & \textbf{88.89}\% & \statpair{0.36}{0.64} \\
\midrule
CORe50 & PPO & \statpair{5.2}{1.7} & \statpair{3.5}{1.2} & \statpair{7.52}{7.12} & \statpair{0.91}{0.04} & \statpair{0.83}{0.06} & \statpair{0.90}{0.04} & 47.37\% & \statpair{0.51}{0.98} \\
 & Pinching & \statpair{2.3}{0.8} & \statpair{6.1}{1.4} & \statpair{4.03}{1.76} & \statpair{0.86}{0.06} & \statpair{0.98}{0.04} & \statpair{\textbf{1.00}}{0.00} & 73.68\% & \statpair{0.28}{0.38} \\
 & \textbf{Ours} & \statpair{\textbf{0.3}}{0.5} & \statpair{\textbf{1.4}}{2.1} & \statpair{\textbf{2.27}}{4.30} & \statpair{\textbf{0.96}}{0.07} & \statpair{\textbf{1.00}}{0.00} & \statpair{\textbf{1.00}}{0.00} & \textbf{78.95}\% & \statpair{\textbf{1.27}}{1.95} \\
\bottomrule
\end{tabular}
\end{adjustbox}
\end{table}

Table~\ref{tab:recon-hand} quantifies how the hand pose changes along the pipeline. Most of the changes occur in the visual warm-up, while the physics-based refinement moves the pose marginally. This is due to the differences among our simulation-ready hand proxy, the reconstructed object, and the dataset annotations used as the reference, so a physically valid grasp need not coincide exactly with the annotated pose. This is a reasonable consequence of pursuing geometric and physical plausibility and stability, while the qualitative results show that our reconstructions remain visually faithful to the input images.

\begin{table}[!t]
\centering
\small
\setlength{\tabcolsep}{2.5pt}
\renewcommand{\arraystretch}{1.08}
\caption{Hand reconstruction quality before optimization (HORT), after visual warm-up, and after the full pipeline (mean $\pm$ standard deviation).}
\label{tab:recon-hand}
\begin{adjustbox}{max width=\linewidth}
\begin{tabular}{llccc}
\toprule
Dataset & Stage & MPJPE-3D (mm) $\downarrow$ & PCK@0.025 $\uparrow$ & PCK@0.05 $\uparrow$ \\
\midrule
DexYCB & HORT & \statpair{7.7}{3.2} & \statpair{0.68}{0.20} & \statpair{0.92}{0.11} \\
 & After warm-up & \statpair{11.0}{3.8} & \statpair{0.62}{0.20} & \statpair{0.91}{0.12} \\
 & Ours & \statpair{11.3}{3.8} & \statpair{0.60}{0.20} & \statpair{0.89}{0.12} \\
\midrule
HO3D & HORT & \statpair{5.6}{3.9} & \statpair{0.72}{0.24} & \statpair{0.94}{0.11} \\
 & After warm-up & \statpair{10.7}{4.5} & \statpair{0.60}{0.22} & \statpair{0.88}{0.14} \\
 & Ours & \statpair{11.0}{4.4} & \statpair{0.59}{0.22} & \statpair{0.87}{0.15} \\
\bottomrule
\end{tabular}
\end{adjustbox}
\end{table}

\FloatBarrier
\subsection{Ablation Studies}

We ablate warm-up on 86 examples (61 DexYCB, 25 HO3D), with equal
maximum budgets: 600 warm-up plus 2000 physics-based iterations,
or 2600 physics-based iterations without warm-up. Both use early stopping as in our experiments in Section~\ref{sec:baseline}. Removing warm-up reduces success
from $90.70\%$ to $76.74\%$ and increases hand motion on both
datasets (Table~\ref{tab:warmup-ablation}), demonstrating the benefit
of contact initialization.

\begin{table}[!htbp]
\centering
\small
\setlength{\tabcolsep}{2.5pt}
\renewcommand{\arraystretch}{1.08}
\caption{Visual warm-up ablation on 86 test cases. Metrics and aggregation follow Table~\ref{tab:quantitative}; values are mean $\pm$ standard deviation.}
\begin{adjustbox}{max width=\linewidth}
\begin{tabular}{llccccccc}
\toprule
Dataset & Method & \shortstack{MPJPE\\(mm) $\downarrow$} & \shortstack{Obj. trans.\\(mm) $\downarrow$} & \shortstack{Obj. rot.\\($^\circ$) $\downarrow$} & \shortstack{Object\\IoU $\uparrow$} & \shortstack{PCK $\uparrow$\\@0.025} & \shortstack{Success\\rate $\uparrow$} & \shortstack{Drop time\\(s) $\uparrow$} \\
\midrule
\multirow{2}{*}{\textbf{DexYCB}} & No warm-up & \statpair{0.4}{0.6} & \statpair{\textbf{1.1}}{1.7} & \statpair{1.15}{2.17} & \statpair{\textbf{0.98}}{0.04} & \statpair{\textbf{1.00}}{0.00} & 78.69\% & \statpair{\textbf{0.07}}{0.05} \\
  & Warm-up & \statpair{\textbf{0.2}}{0.2} & \statpair{\textbf{1.1}}{2.1} & \statpair{\textbf{0.94}}{1.56} & \statpair{0.97}{0.04} & \statpair{\textbf{1.00}}{0.00} & \textbf{91.80}\% & \statpair{0.06}{0.01} \\
\midrule
\multirow{2}{*}{\textbf{HO3D}} & No warm-up & \statpair{0.3}{0.3} & \statpair{\textbf{1.1}}{1.9} & \statpair{\textbf{0.88}}{1.47} & \statpair{\textbf{0.98}}{0.05} & \statpair{\textbf{1.00}}{0.00} & 72.00\% & \statpair{0.04}{0.01} \\
  & Warm-up & \statpair{\textbf{0.2}}{0.3} & \statpair{1.3}{1.4} & \statpair{0.93}{1.02} & \statpair{0.97}{0.04} & \statpair{\textbf{1.00}}{0.00} & \textbf{88.00}\% & \statpair{\textbf{0.05}}{0.00} \\
\bottomrule
\end{tabular}
\end{adjustbox}
\label{tab:warmup-ablation}
\end{table}

\begin{figure}[!htbp]
\centering
\begingroup
\setlength{\tabcolsep}{0pt}
\renewcommand{\arraystretch}{1}
\begin{minipage}[c]{0.135\linewidth}
  \centering\footnotesize
  \textbf{Reference}\par\vspace{3pt}
  Input image\par\vspace{1pt}
  \includegraphics[width=0.96\linewidth,height=0.96\linewidth,keepaspectratio]{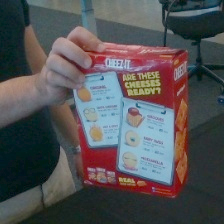}\par
  \vspace{4pt}
  HORT initialization\par\vspace{1pt}
  \includegraphics[width=0.96\linewidth,height=0.96\linewidth,keepaspectratio]{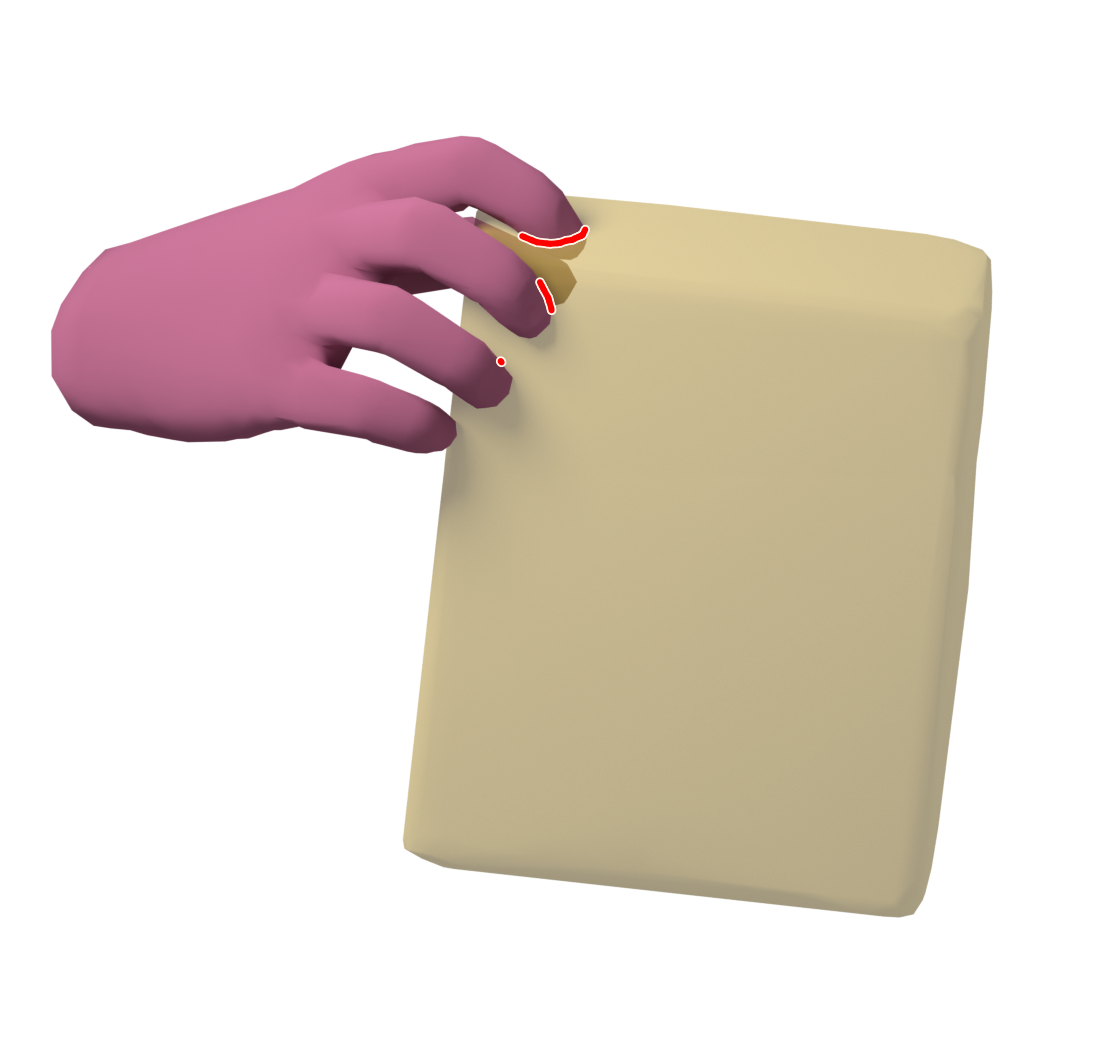}
\end{minipage}%
\hspace{0.015\linewidth}%
\raisebox{-0.5\height}{\rule{0.3pt}{0.33\linewidth}}%
\hspace{0.015\linewidth}%
\begin{minipage}[c]{0.825\linewidth}
\centering\footnotesize
\textbf{Visual warm-up ablations}\par\vspace{3pt}
\newcommand{\ablone}[1]{\includegraphics[width=0.18\linewidth,height=0.15\linewidth,keepaspectratio]{#1}}
\newcommand{\abltwo}[1]{\includegraphics[width=0.39\linewidth,height=0.15\linewidth,keepaspectratio]{#1}}
\begin{tabular}{@{}c@{\hspace{0.01\linewidth}}c@{\hspace{0.01\linewidth}}c@{\hspace{0.01\linewidth}}c@{}}
With $\mathcal{L}_{\mathrm{obs}}$ & With $\mathcal{L}_{\mathrm{lim}}$ & With $\mathcal{L}_{\mathrm{int}}$ & With $\mathcal{L}_{\mathrm{adh}}$\\[1pt]
\ablone{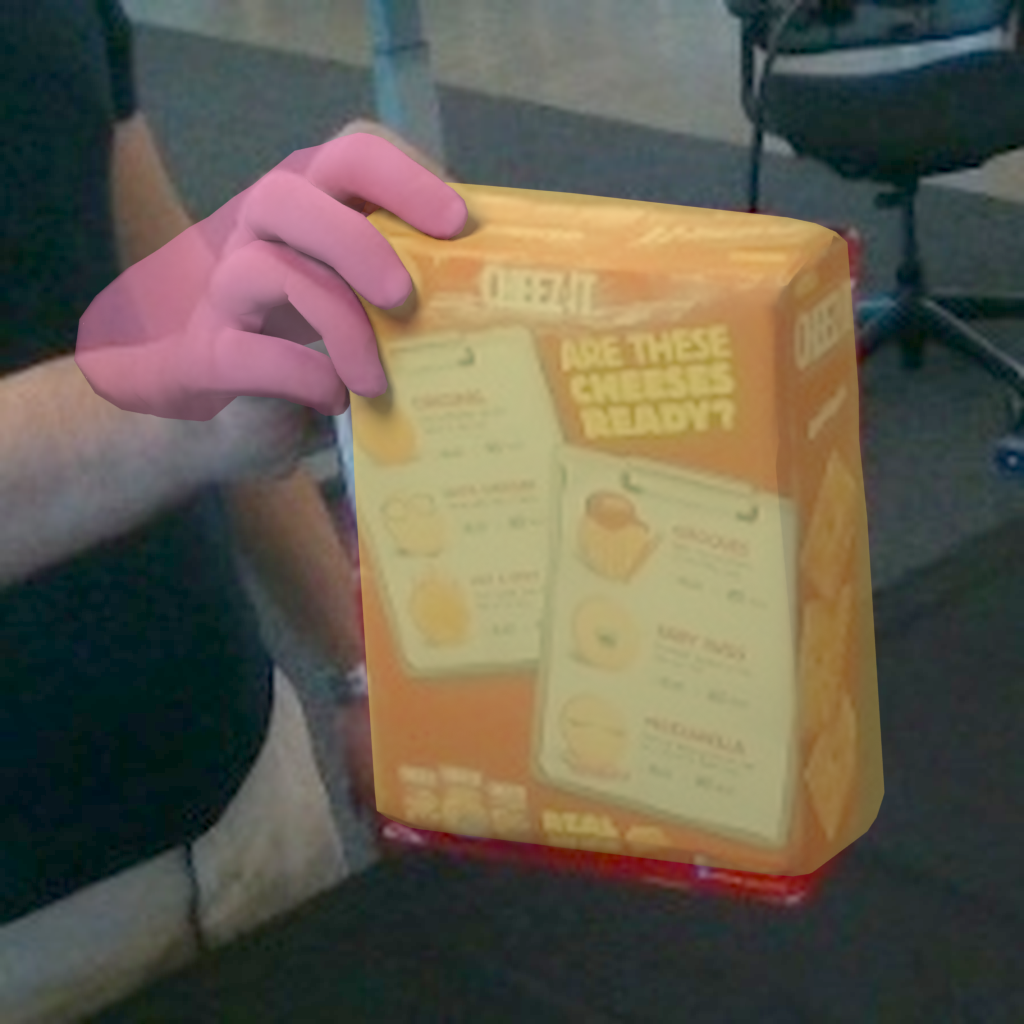} & \ablone{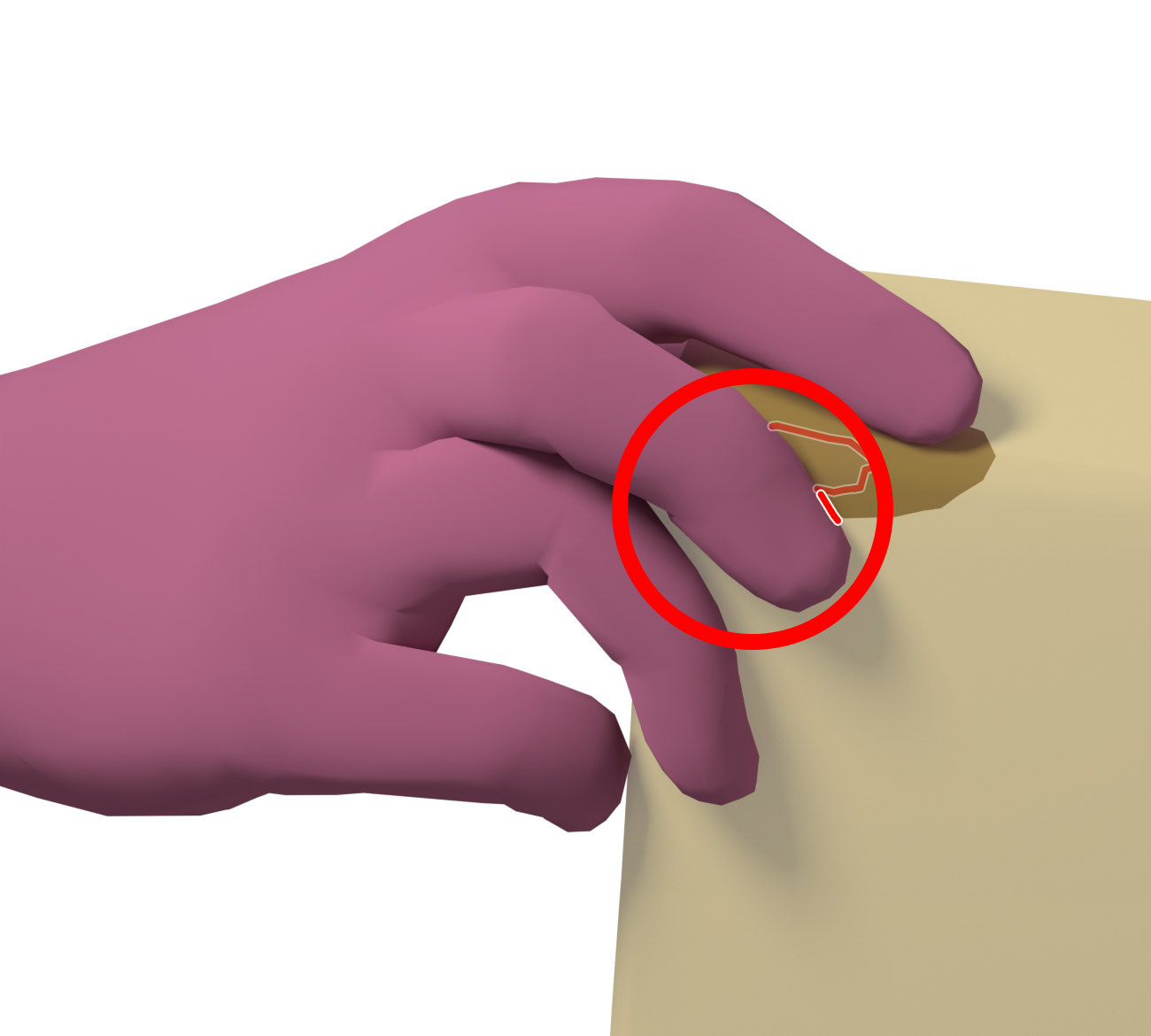} & \ablone{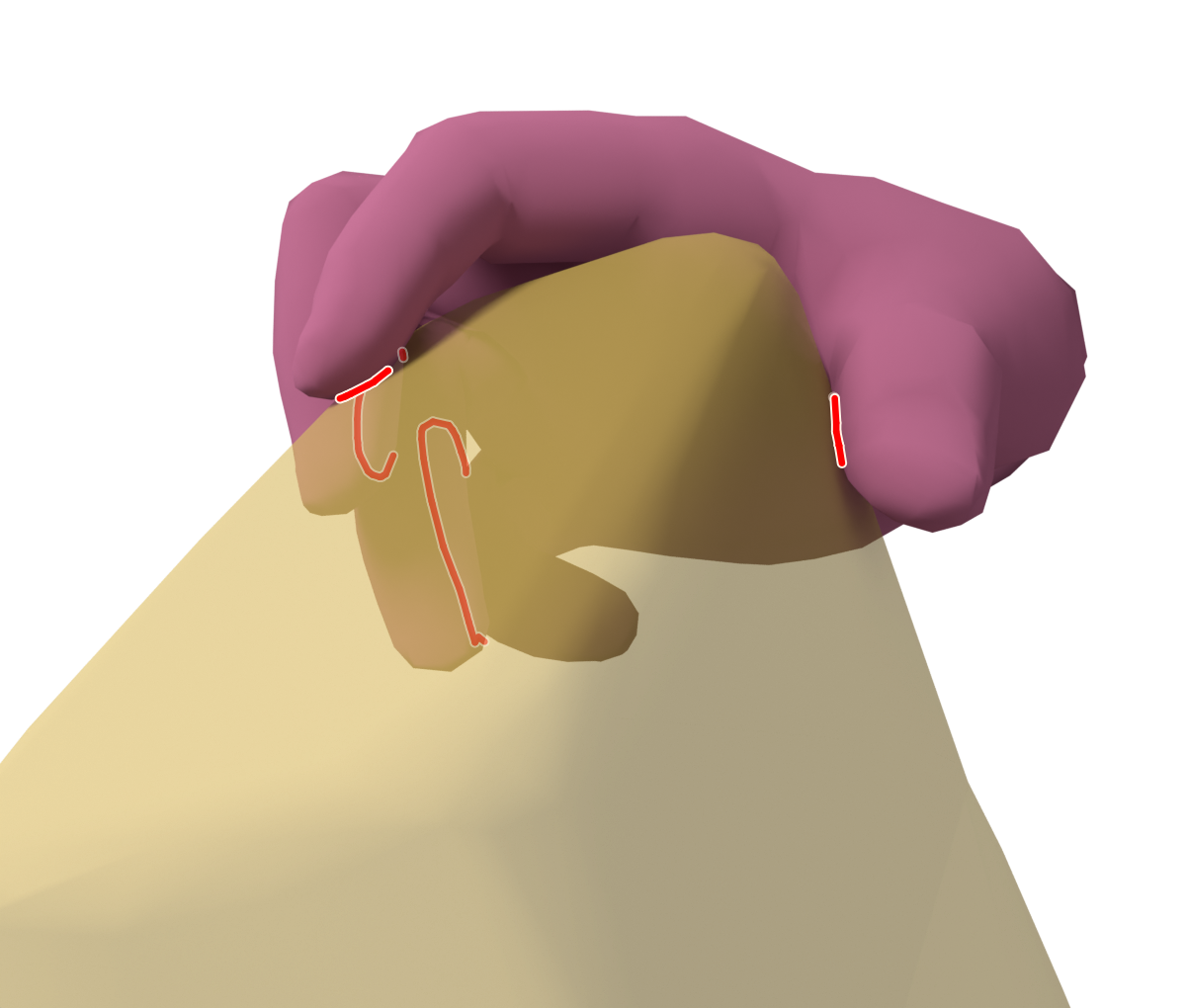} & \abltwo{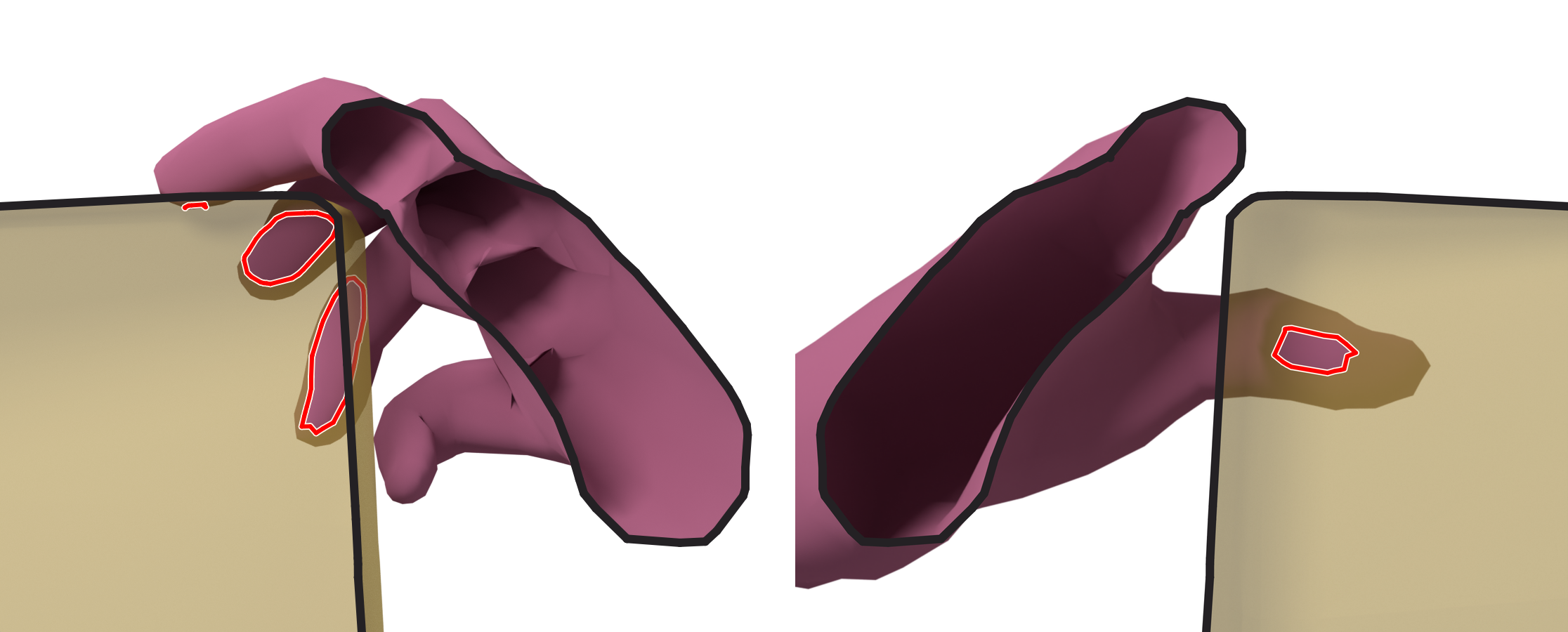}\\[4pt]
Without $\mathcal{L}_{\mathrm{obs}}$ & Without $\mathcal{L}_{\mathrm{lim}}$ & Without $\mathcal{L}_{\mathrm{int}}$ & Without $\mathcal{L}_{\mathrm{adh}}$\\[1pt]
\ablone{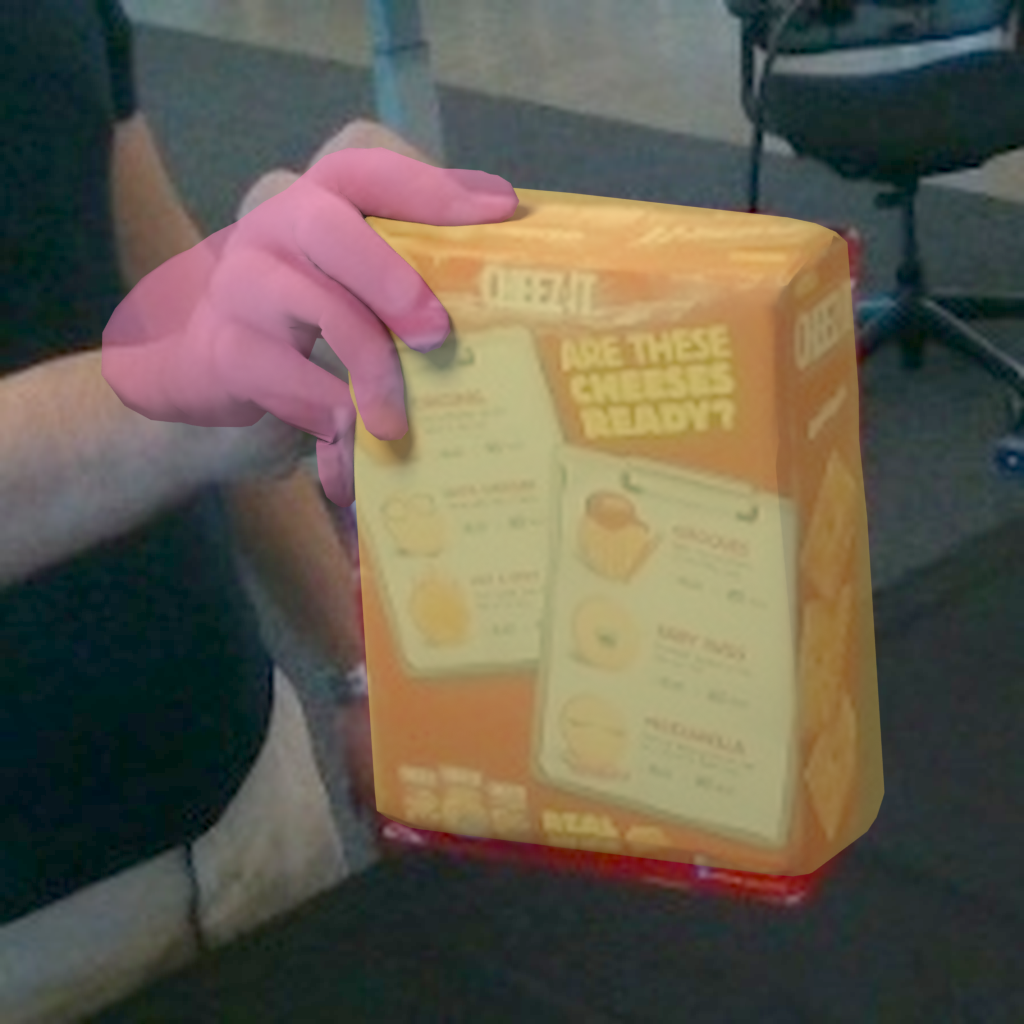} & \ablone{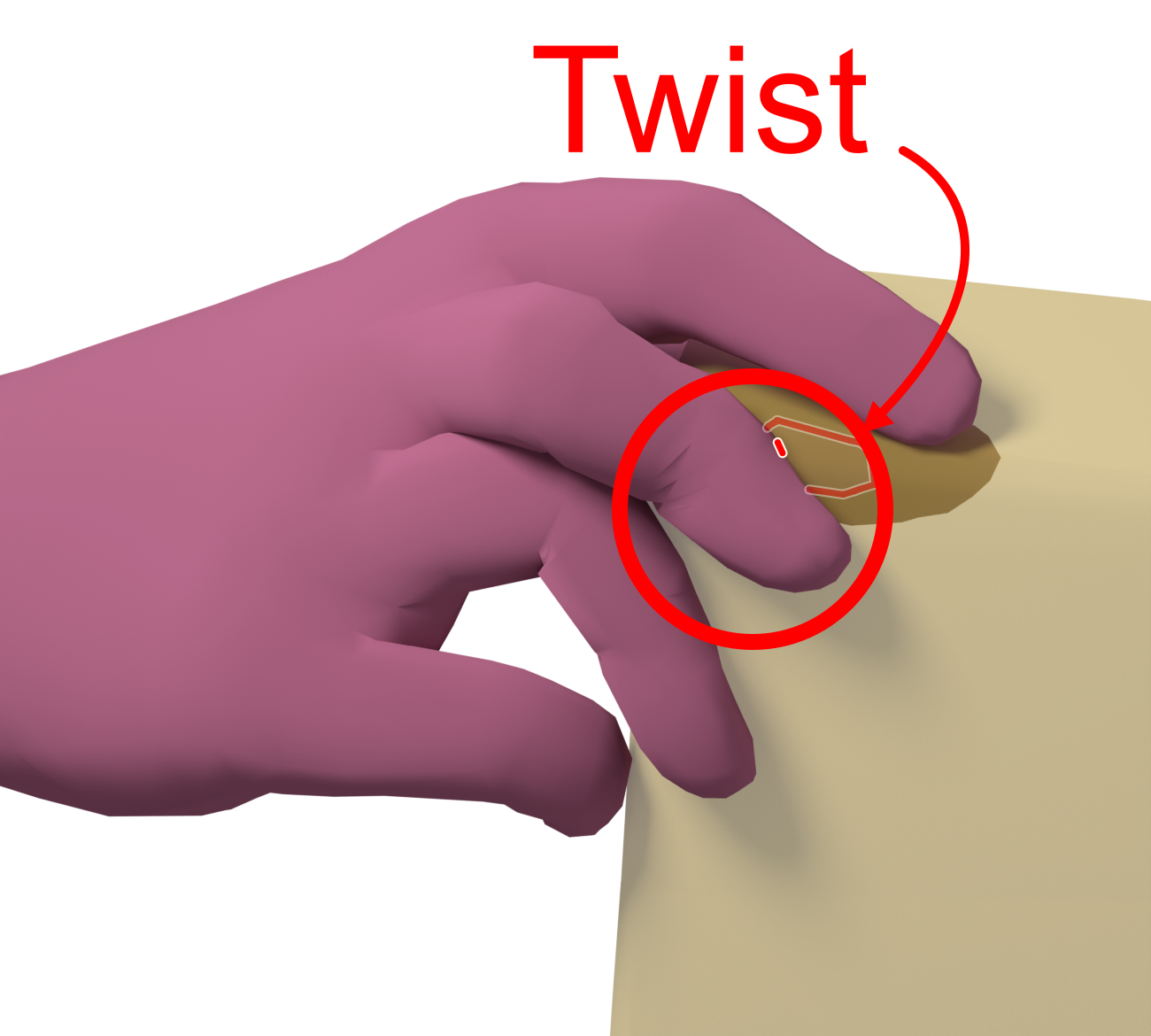} & \ablone{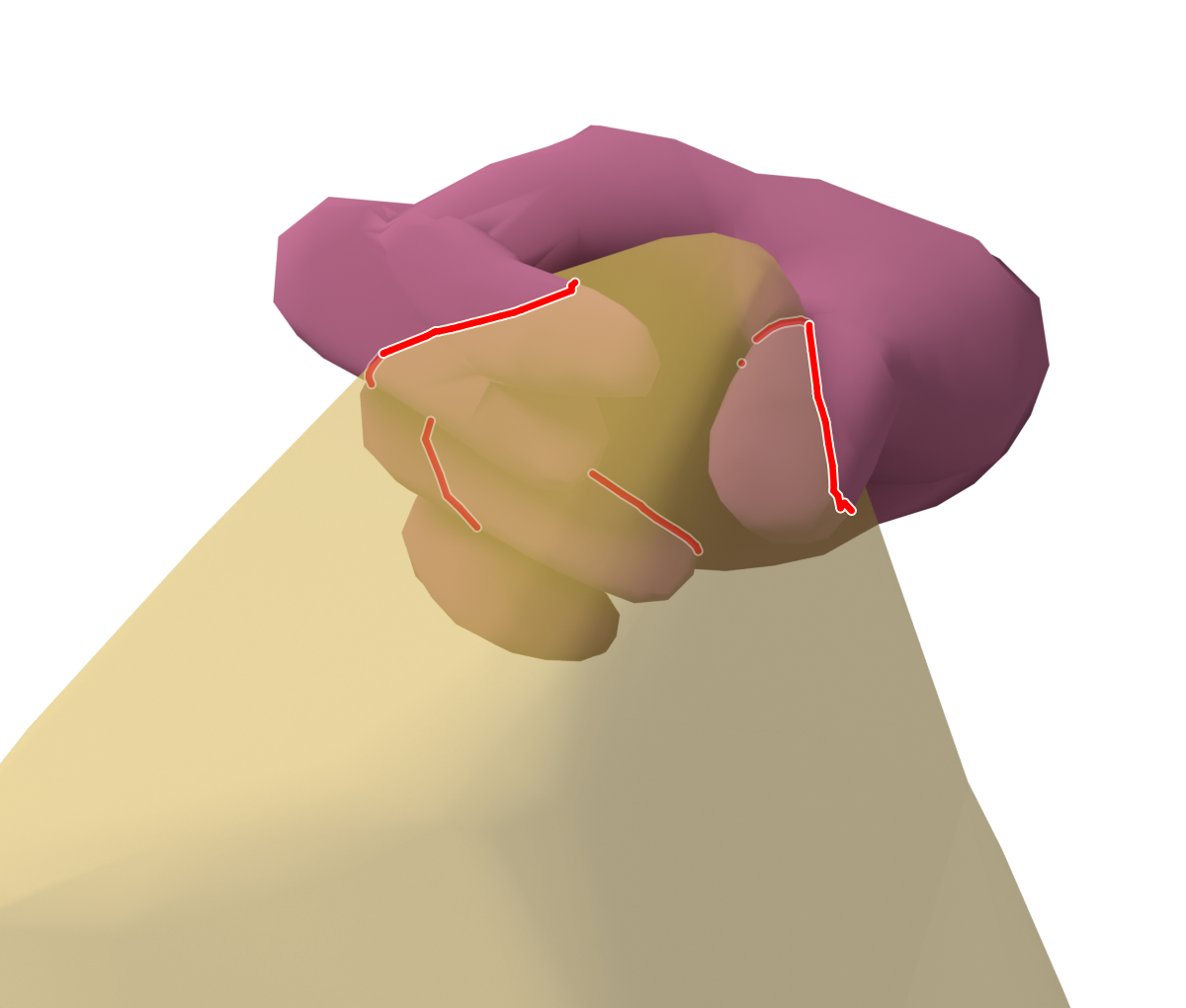} & \abltwo{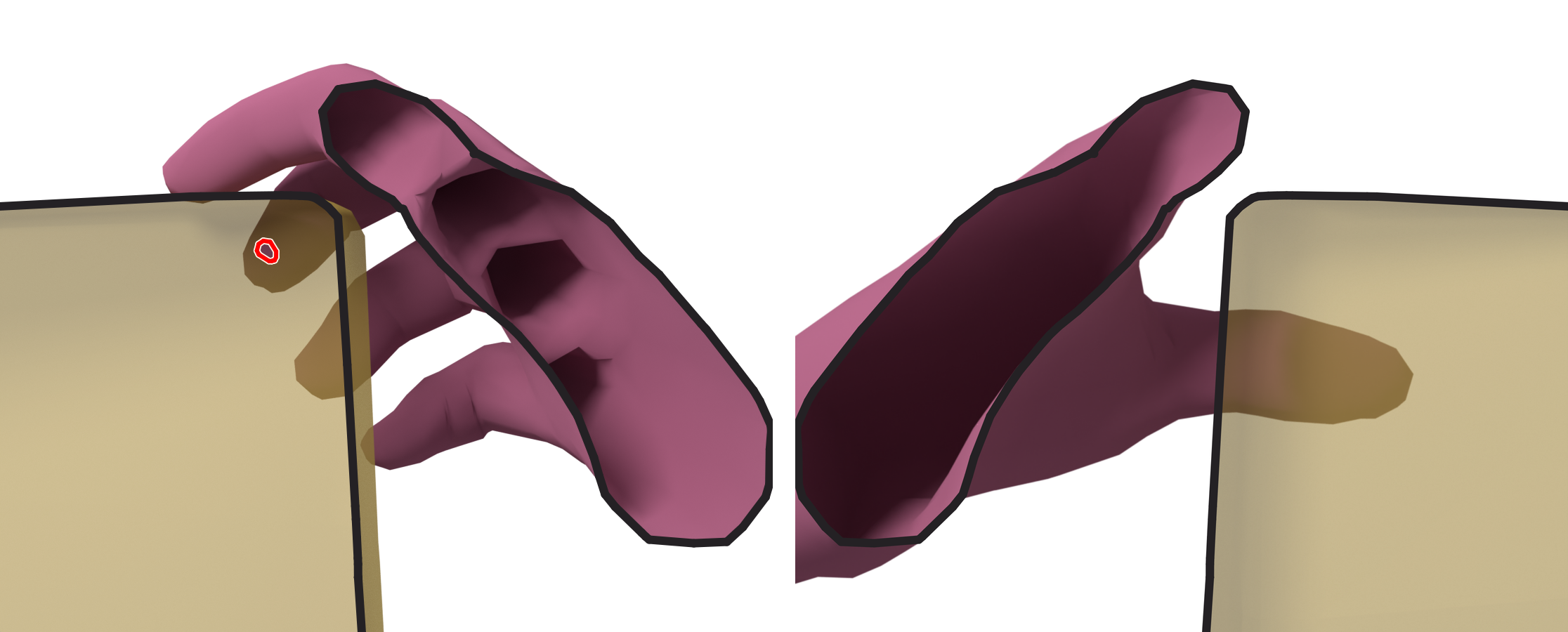}
\end{tabular}
\end{minipage}
\endgroup
\caption{Visual warm-up ablations. Left: input and HORT initialization as references. Right: complete warm-up (top) and individual regularizer ablations (bottom), with matching views in each column. Without $\mathcal{L}_{\mathrm{obs}}$, the reconstructed grasp drifts from the input image. Without $\mathcal{L}_{\mathrm{lim}}$, the middle finger's distal joint exhibits severe, biomechanically implausible twisting. Without $\mathcal{L}_{\mathrm{int}}$, the hand penetrates deeply into the object. Without $\mathcal{L}_{\mathrm{adh}}$, the fingers make insufficient contact with the object.}
\label{fig:ablations}
\end{figure}

Figure~\ref{fig:ablations} further shows how each warm-up
regularizer contributes to a plausible initial grasp.

\FloatBarrier
\section{Conclusion}
\label{sec:5}

We presented StableGrasp, a method for reconstructing stable human
grasps from single RGB images. By jointly optimizing the visual hand
pose and a PD control target, it minimizes the kinetic energy of the
coupled hand--object system while enforcing visual and geometric
constraints. Experiments demonstrate improved grasp stability over baselines while preserving agreement with the input image.

\paragraph{Limitations and Future Work}
Our local pose optimization keeps object geometry fixed and therefore
depends on the initial reconstruction. Alpha-shape meshing can distort
thin objects and fill concave or hollow features. Object reconstruction
errors account for 27 out of all our 48 failure cases (on the 362 examples we tested). The rigid hand proxy
also omits local tissue deformation, which affects contact geometry
and force transmission. Future work could combine improved object
reconstruction with deformable hand models.

\clearpage

\bibliographystyle{iclr2027_conference}
\bibliography{paper}

\end{document}